\documentclass[10pt,twocolumn,letterpaper]{article}

\usepackage{iccv}
\usepackage{times}
\usepackage{epsfig}
\usepackage{graphicx}
\usepackage{amsmath}
\usepackage{amssymb}
\usepackage{balance}
\usepackage{multirow}

\usepackage[pagebackref=true,breaklinks=true,letterpaper=true,colorlinks,bookmarks=false]{hyperref}

\newcommand{\R}{\mathbb{R}}

\iccvfinalcopy 

\def\iccvPaperID{14} 

\ificcvfinal\fi

\begin{document}

\title{Audio-Visual Transformer Based Crowd Counting}

\author{Usman Sajid$^1$, Xiangyu Chen$^1$, Hasan Sajid$^2$, Taejoon Kim$^1$, Guanghui Wang$^3$\\
$^1$\textit{Electrical Engineering and Computer Science,} 
\textit{University of Kansas,
Lawrence, KS, USA, 66045} \\
$^2$ \textit{School of Mechanical and Manufacturing Engineering, NUST, Islamabad, Pakistan}\\
$^3$ \textit{Department of Computer Science},
\textit{Ryerson University,
Toronto, ON, Canada M5B 2K3}\\
{\tt\small Email: \{usajid,xychen,taejoonkim\}@ku.edu$^1$, hasan.sajid@smme.nust.edu.pk$^2$, wangcs@ryerson.ca$^3$}

}

\maketitle
\ificcvfinal\thispagestyle{empty}\fi

\begin{abstract}
Crowd estimation is a very challenging problem. The most recent study tries to exploit auditory information to aid the visual models, however, the performance is limited due to the lack of an effective approach for feature extraction and integration. The paper proposes a new audio-visual multi-task network to address the critical challenges in crowd counting by effectively utilizing both visual and audio inputs for better modalities association and productive feature extraction. The proposed network introduces the notion of auxiliary and explicit image patch-importance ranking (PIR) and patch-wise crowd estimate (PCE) information to produce a third (run-time) modality. These modalities (audio, visual, run-time) undergo a transformer-inspired cross-modality co-attention mechanism to finally output the crowd estimate. To acquire rich visual features, we propose a multi-branch structure with transformer-style fusion in-between. Extensive experimental evaluations show that the proposed scheme outperforms the state-of-the-art networks under all evaluation settings with up to $33.8\%$ improvement. We also analyze and compare the vision-only variant of our network and empirically demonstrate its superiority over previous approaches.
\end{abstract}

\section{Introduction}
\begin{figure}[t]
\label{fig:fig1}
\begin{minipage}[b]{0.162\columnwidth}
		\begin{center}
		\end{center}
	\end{minipage}
	\begin{minipage}[b]{0.325\columnwidth}
		\begin{center}
			\centerline{\includegraphics[width=1\columnwidth]{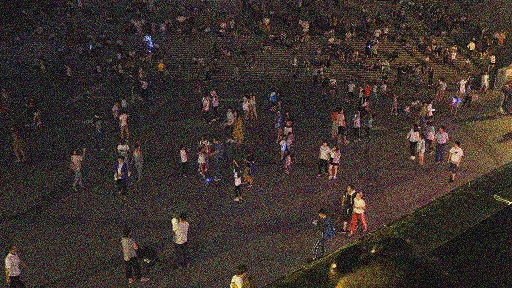}}
			\centerline{\footnotesize{Input Image}}
		\end{center}
	\end{minipage}
	\begin{minipage}[b]{0.325\columnwidth}
		\begin{center}
			\centerline{\includegraphics[width=1\columnwidth]{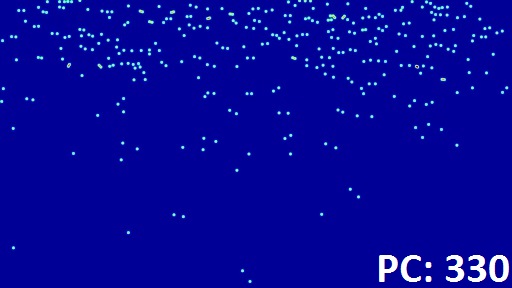}}
			\centerline{\footnotesize{Ground-truth density-map}}
		\end{center}
	\end{minipage}
	\begin{minipage}[b]{0.162\columnwidth}
		\begin{center}
		\end{center}
	\end{minipage}
		\begin{minipage}[l]{0.325\columnwidth}
		\begin{center}
			\centerline{\includegraphics[width=1.0\columnwidth]{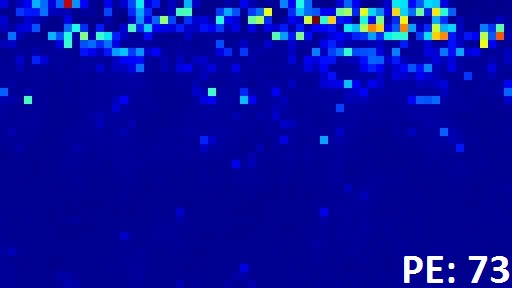}}
			\centerline{\footnotesize{Visual-only \cite{hu2020ambient}}}
		    \end{center}
	\end{minipage}
	\begin{minipage}[r]{0.325\columnwidth}
		\begin{center}
			\centerline{\includegraphics[width=1.0\columnwidth]{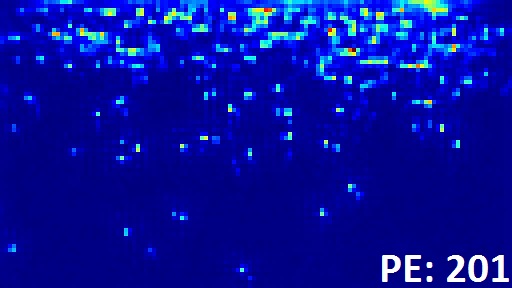}}
			\centerline{\footnotesize{Audio-Visual \cite{hu2020ambient}}}
		\end{center}
	\end{minipage}
	\begin{minipage}[r]{0.325\columnwidth}
		\begin{center}
			\centerline{\includegraphics[width=1.0\columnwidth]{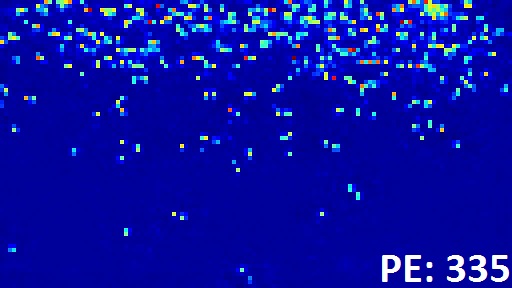}}
			\centerline{\footnotesize{Audio-Visual (Ours)}}
		\end{center}
	\end{minipage}
    \vspace{-1mm}
	\caption{\footnotesize{
	For the low-quality input image with severe conditions such as noise, low-illumination, or low-resolution, the proposed audio-visual model yields the best and more fine-grained people estimate (PE) as evaluated using the ground-truth density-map and people count (PC).
	}}
    \vspace{-3mm}
    
\end{figure}

Crowd estimation requires one to count the total people in the given image. It finds many applications in real-world scenarios, e.g., better management of crowd gatherings, safety and security, and circumventing any undesirable incident. Many deep learning-based image-only schemes \cite{sajid2020multi,sam2017switching,sajid2020plug,idrees2018composition,sajid2020plug,jiang2019crowd,zhang2016single,liu2018leveraging} have been proposed to date, ranging from single and multi-branch networks \cite{zhang2016single,sajid2020multi,sajid2020plug}, multi-regressors \cite{sam2017switching} based to trellis networks \cite{jiang2019crowd}. Although they show reasonable performance in regular images, they fail to generalize well in many practical scenarios such as low illumination and lighting conditions, noise, severe occlusion, and low-resolution images, where visual information is scarce. Consequently, they give huge crowd under-estimation as shown in Fig. \ref{fig:fig1}. Lack of visual clues may also invoke highly sensitized behavior in these models towards different image regions, resulting in large over-estimation. Moreover, in the case of regular images, sub-optimal capabilities of these state-of-the-arts implicate that there is a lot of room for further improvement.

One compelling way to address these challenges is to investigate the effect of utilizing more than one modality (e.g., image and audio). Recently, Hu \textit{et al.} \cite{hu2020ambient} introduced a novel audio-visual crowd counting task and proposed an estimation model that jointly learns both visual and audio features and fuses them together. The results demonstrate that combining the related audio modality with the visual input significantly improves the crowd estimate in such conditions. 
However, it only accounts for the parametric influence of audio features on the visual ones without making full use of the audio-visual information, thus, under- or over-estimating the crowd as shown in Fig. \ref{fig:fig1}.

On the other hand, the learning and fusion of visual and audio modalities have been applied with reasonable success to other computer vision problems, e.g. classification tasks \cite{wu2014exploring,cen2021deep, hu2016temporal,kiela2018efficient}, event localization \cite{lin2020audiovisual,xuan2020cross}, and speech recognition \cite{yuhas1989integration,gurban2008dynamic,mroueh2015deep,sterpu2018attention}. However, these schemes are generally not suitable for the crowd estimation task because of very few pixels per person, and thus require a specifically tailored method to obtain pixel-perfect results. Moreover, these schemes (including \cite{hu2020ambient}) mostly focus on improving the intra- or inter-modality fusion process, and often ignore the significant visual feature extraction part by normally using the conventional VGG \cite{simonyan2014very} or ResNet-based \cite{he2016deep} standard structures for that.

\begin{figure}[t]
	\begin{minipage}[b]{1.0\columnwidth}
		\begin{center}
			\centerline{\includegraphics[width=0.75\columnwidth]{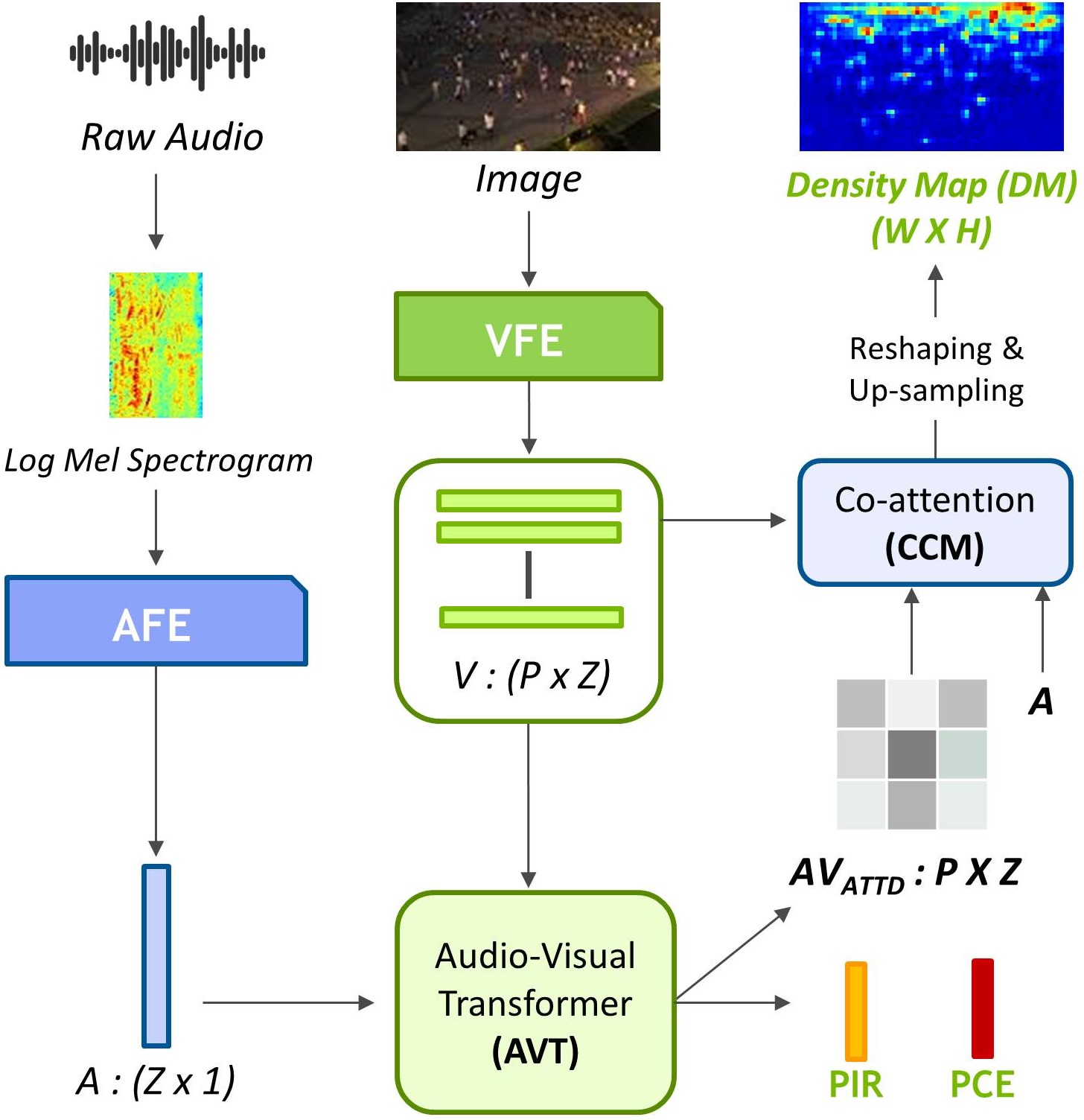}}
		\end{center}
	\end{minipage}		
    \vspace{-4mm}
	\caption{The proposed audio-visual crowd counting network. The extracted audio-visual features ($V,A$) go through the $AVT$ module to obtain two auxiliary network outputs ($PIR,PCE$) and third (run-time) modality ($AV_{ATTD}$). The $AV_{ATTD}$ undergoes the cross-modality co-attention fusion with $V$ and $A$ via the $CCM$ module, followed by getting the final crowd density-map ($DM$).}
    \vspace{-2mm}
    \label{fig:fig2_architecture}
\end{figure}

To address these major challenges and issues, we propose a new transformer-based \cite{vaswani2017attention} audio-visual multi-task crowd counting network as shown in Fig. \ref{fig:fig2_architecture}. It consists of an Audio-Visual Transformer (AVT) that generates two auxiliary network outputs, image patch-importance ranking (PIR) and patch-wise crowd estimate (PCE), as part of the inter-modality fusion process. This explicit PIR and PCE information also helps AVT module in generating a third run-time audio-visual attended modality that consequently helps in constructive association and transformer-style co-attention of audio-visual features. Furthermore, no extra ground-truth annotation process is required to embed the PIR and PCE into the proposed network. Second, instead of deploying the conventional and standard structure for visual feature extraction, we use the multi-scale branches that also undergo the unique transformer-inspired inter-scale fusion process to yield rich and productive visual representations. Extensive experiments show that the proposed model outperforms the state-of-the-art methods in all settings with up to $33.8\%$ improvement, especially in challenging situations such as shown in Fig. \ref{fig:fig1}. 
The main contributions of our work include:\\[-14pt]
\begin{itemize}\setlength\itemsep{-1.5em}
  \item We propose a novel audio-visual multi-task crowd counting network for effective estimation in both regular and severe conditions. To the best of our knowledge, this is the first attempt to use the transformer-style mechanism for this task.
\\
 \item We introduce the notion of auxiliary PIR and PCE information, and empirically shown that it is beneficial for better modalities association and extracting rich visual features without requiring any extra ground-truth annotation process. 
\\
\item We also design an image-only variant of our model. Extensive experimental evaluations on benchmark datasets indicate that the proposed networks significantly outperform the state-of-the-art. The source code will be available at \url{https://github.com/rucv/avcc}.
\end{itemize}

\section{Related Work}

\noindent \textbf{Audio-Visual Learning.} Audio-visual representation learning aims to aid the visual modality with audio or vice-versa. Early speech perception research \cite{mcgurk1976hearing} demonstrates that the visual information can change what people hear, i.e., McGurk Effect. Since then, vision and audio modalities are widely explored in speech recognition\cite{yuhas1989integration,gurban2008dynamic,mroueh2015deep,sterpu2018attention}, video classification \cite{wu2014exploring}, emotion recognition \cite{kim2013deep} and video description \cite{jin2016video}. Multiple kernels are broadly implemented as the fusion module by feeding the kernels with data from different modalities \cite{chen2014emotion, sikka2013multiple, wei2014multimodal}. Another fusion method is based on graphical models considering its advantages in temporal related tasks \cite{gurban2008dynamic, hershey2004audio}. Besides, neural networks raise more attention in fusion especially since the appearance of RNN and LSTM \cite{petridis2017end, sterpu2018attention}. More recently, transformer-based \cite{vaswani2017attention} fusion raises growing attention \cite{alberti2019fusion, sun2019videobert, rahman2019integrating, huang2020multimodal, kan2020metric}, especially after its application in vision \cite{dosovitskiy2020image}. In addition to that, there are also some model-agnostic fusion methods, including the simple concatenation \cite{li2020deep,chen2021few,yu2020audio} and element-wise operation \cite{gan2020music, tian2018audio}.

\begin{figure*}
	\begin{minipage}[b]{1.0\textwidth}
		\begin{center}
			\centerline{\includegraphics[width=0.79\textwidth]{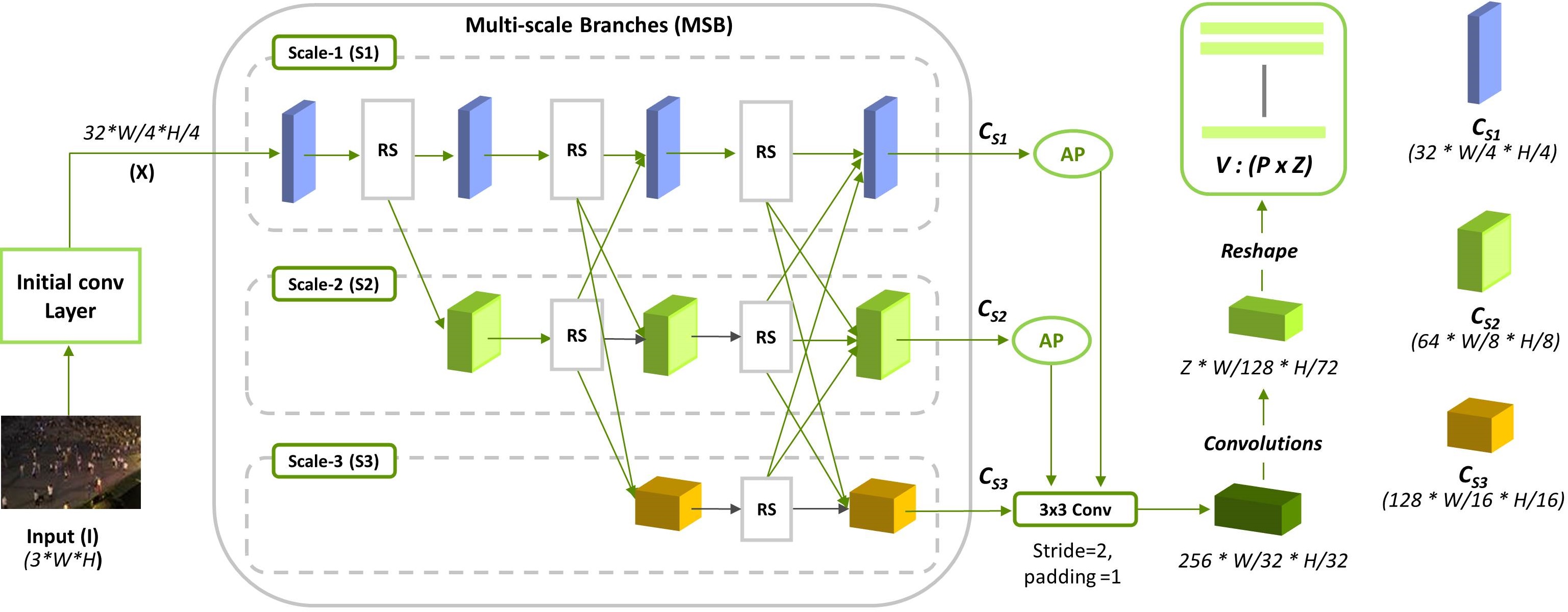}}
		\end{center}
	\end{minipage}
    \vspace{-6mm}
	\caption{The framework of Visual Feature Extraction (VFE) block.}
    \vspace{-2mm}
    \label{fig:fig2_vfe}
\end{figure*}

\noindent \textbf{Crowd Counting.} The research of people count mainly focuses on image-only crowd estimation, and targets several issues such as varying crowd-density and scale, large perspective and heavy occlusions. They are of three categories: Count-by-detection (Det), by-direct-regression (DReg), and by-density-map (DMap). The Det methods \cite{shami2018people,li2019headnet} detect each person via some standard object detectors (e.g. Faster-RCNN \cite{girshick2015fast}, YOLO \cite{redmon2016you}). These methods give unsatisfactory results in the high-density crowd scenarios. The DReg models \cite{sajid2020multi,sajid2020plug,wang2015deep,sajid2020zoomcount,wang2015deep} directly regress the crowd number using CNN-based structures. Wang \textit{et al.} \cite{wang2015deep} deployed the AlexNet \cite{krizhevsky2012imagenet} variant for direct crowd regression. Recently, Sajid \textit{et al.} designed two different types of direct-regression counting methods \cite{sajid2020plug,sajid2020zoomcount} that use the patch-rescaling module (PRM) and branch structure to deal with varying crowd levels. But these models fail to utilize the valuable density-map based computation. The DMap methods \cite{idrees2018composition,zhang2016single,sam2017switching,liu2018leveraging,sindagi2019ha} estimate crowd density-map, where each pixel indicates crowd-density. The pixel values are then summed-up to obtain final count. Switch-CNN \cite{sam2017switching} uses CCN-based switch that routes the image to one of three specialized regressors, each dealing with specific crowd-density. Li \textit{et al.} \cite{li2018csrnet} employed dilation layers for better contextual information retrieval. Liu \textit{et al.} \cite{liu2018leveraging} used rank-based system for unsupervised learning.  Idrees \textit{et al.} \cite{idrees2018composition} deployed composition loss to jointly learn the count, localization and density-map. HA-CCN \cite{sindagi2019ha} utilized global and spatial attention to enhance useful features. However, these schemes prove inadequate to handle extreme conditions such as noise, low-illumination and resolution images.

In the audio-visual domain, Hu \textit{et al.} \cite{hu2020ambient} recently introduced the first-ever audio-visual crowd dataset, DISCO, making this type of crowd counting possible. For the audio-visual people count, how to constructively extract the audio-visual features and how to effectively fuse them together present the key challenges. Therefore, the proposed DMap-based network focuses on solving these major challenges amid dealing with severe conditions as discussed above.

\section{Proposed Approach}
The proposed multi-task model, as shown in Fig. \ref{fig:fig2_architecture}, exploits both input image and audio modalities for effective crowd estimation. First, we separately extract rich features for both modalities, then pass them through the Audio-Visual Transformer (AVT) to generate two auxiliary network outputs: Patch-Importance Ranking (PIR) and Patch-wise Crowd Estimate (PCE). The explicit PIR and PCE vectors play a  crucial role in improving the final crowd estimate, and also help the AVT in generating the audio-visually attended channels. These attended channels then undergo the cross-modality co-attention process along with the original audio-visual features ($V,A$) via the CCM module. Finally, the CCM output goes through the reshaping and up-sampling steps to give the crowd-density map, where we sum-up all its pixel values to yield the final crowd count. The network components are detailed below.

\subsection{Audio Feature Extraction (AFE)} 
To extract the audio features embedding, we deploy the ResNet-like CNN structure \cite{hershey2017cnn} (pretrained on the AudioSet dataset \cite{kong2018audio}) and apply it on the conventionally computed \cite{hu2020ambient} Log Mel-Spectogram (LMS) representation of the raw one-second duration input audio signal. For the given Audio LMS ($A_{LMS}\in \R^{64*96})$, audio CNN ($AFE$) yields the vector output as follows:
\vspace{0mm}
\begin{equation}
\label{eq2}
A=AFE(A_{LMS})
\end{equation}
where $A\in \R^{Z*1}$ represents the extracted audio embedding.

\subsection{Visual Feature Extraction (VFE)} 
The VFE component, as shown in Fig. \ref{fig:fig2_vfe}, comprises of three multi-scale branches (MSB) with repeated inter-branch fusion. The input image (I $\in \R^{3*W*H}$) passes through two initial ($3\times3$) convolutional layers to obtain the down-scaled channels ($X\in \R^{32*\frac{W}{4}*\frac{H}{4}}$). These features then proceed through the multi-scale branches ($S1,S2,S3$) that are composed of several residual structures (RS). The RS block contains four residual units, where each unit is composed of a three-layer based ResNet building block \cite{he2016deep}. Similar to the high-resolution networks \cite{sun2019deep,wang2020deep}, each branch retains its channel quantity and resolution throughout that branch. Channel quantity doubles each time as we move from $S_1$ to $S_3$, while the resolution decreases by half. The MSB module outputs three separate sets of channels ($C_{S1}\in \R^{32*\frac{W}{4}*\frac{H}{4}},C_{S2}\in \R^{64*\frac{W}{8}*\frac{H}{8}},C_{S3}\in \R^{128*\frac{W}{16}*\frac{H}{16}}$).

\begin{figure}[t]
	\begin{minipage}[b]{1\columnwidth}
		\begin{center}
			\centerline{\includegraphics[width=0.90\columnwidth]{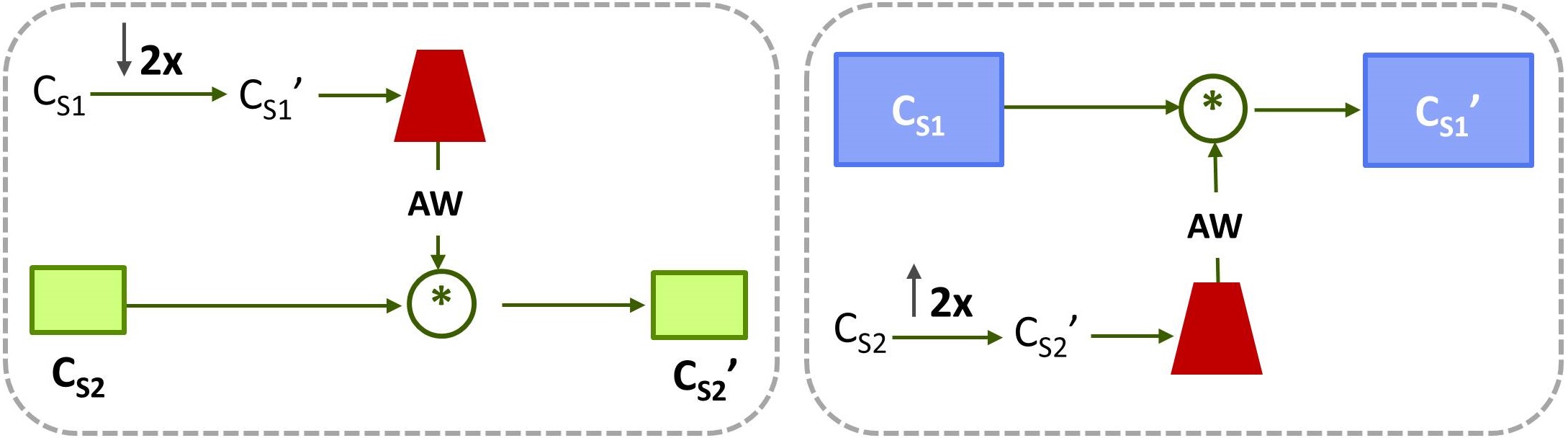}}
		\end{center}
	\end{minipage}
   \begin{minipage}[b]{1\columnwidth}
		\begin{center}
			\centerline{\includegraphics[width=0.88\columnwidth]{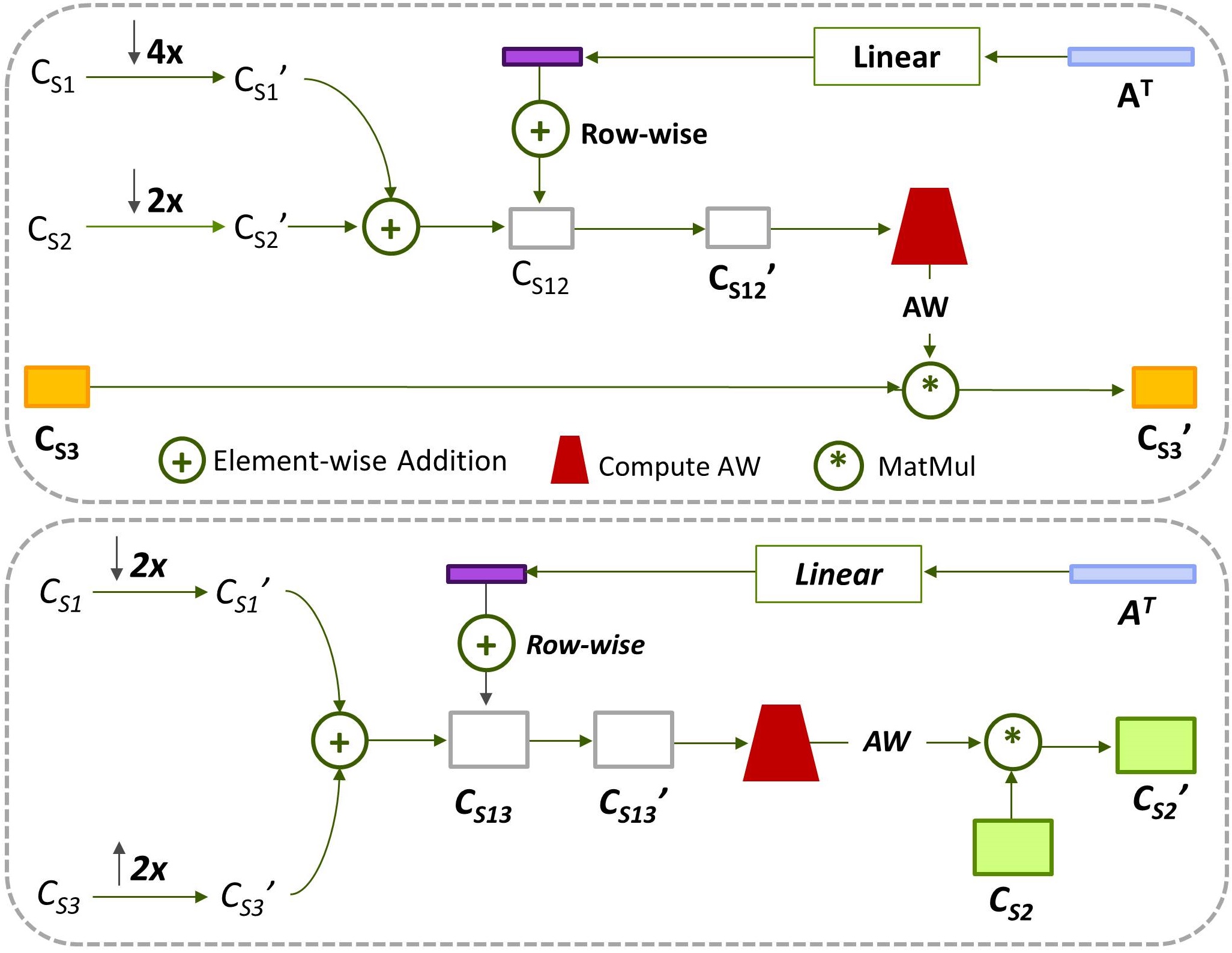}}
			
		\end{center}
	\end{minipage}
	\vspace{-6mm}
	\caption{\footnotesize{Illustration of different transformer-inspired inter-branch fusion cases. (S1 $\rightarrow$ S2 only (top-left), S2 $\rightarrow$ S1 only (top-right), \{S1, S2\} $\rightarrow$ S3 (middle), \{S1, S3\}, $\rightarrow$ S2 (bottom))
	}}
    \vspace{-1mm}
    \label{fig:fusion_tech}
\end{figure}

\subsubsection{Inter-Branch Fusion} 
The purpose of inter-branch fusion is to develop coordinated knowledge in-between the multi-scale branches. We denote this process as $\{S\}\rightarrow T$, indicating the fusion from one or two source branches ($S$) channels into the target branch ($T$) features. We deploy the transformer-inspired attention mechanism to achieve such fusion. All cases are detailed below as well as illustrated in Fig. \ref{fig:fusion_tech}. It is worth mentioning that later stage three-branch fusions also integrate the audio embedding ($A$) during the fusion process, which empirically proves beneficial and also unique multi-modal strategy to the proposed method. The fusion process significantly helps the VFE in preparing constructive and co-attended visual features ($V$) for the next steps. 

\noindent \textbf{S1$\rightarrow$ S2 only (and vice versa).} In this step, we first down-sample the source branch channels ($C_{S1}$) via $3\times3$ convolution to match the resolution and quantity of the S2 channels. The resultant channels ($C_{S1}'$) are converted into attention-weights (AW), which separately undergo the attention mechanism with the respective target branch channels ($C_{S2}$) to give visually-attended features ($C_{S2}'$) as shown in Fig. \ref{fig:fusion_tech}. Mathematically, it is defined as:
\begin{equation}
\label{eq3}
C_{S2}' = AW*C_{S2}=softmax(C_{S1}' * C_{S1}^{'T})*C_{S2}
\end{equation}
where $*$ and $T$ denote matrix multiplication (MatMul) and transpose respectively. In case of \textbf{(S2$\rightarrow$ S1 only) fusion}, the approach remains same except that the lower-branch channels ($C_{S2}$) are bi-linearly up-sampled to match $C_{S1}$ features dimensions before fusing together as shown in Fig. \ref{fig:fusion_tech}.

\noindent \textbf{\{S1, S2\}$\rightarrow$ S3 Fusion case.} 
Both higher-branch source channels ($C_{S1},C_{S2}$) get down-scaled to match the lowest-branch channels ($C_{S3}$) dimensions. The generated channels ($C_{S1}'$,$C_{S2}'$) are added element-wise to produce features $C_{S12}$. After the linear-layer operation on the audio embedding row-vector ($A^T$), it separately performs element-wise addition with each row  of $C_{S12}$. The resultant $C_{S12}'$ is being used next to obtain the attention weights (AW). The AW finally gets applied on the target branch channels ($C_{S3}$) to produce the audio-visual attended channels ($C_{S3}'$) as shown in Fig. \ref{fig:fusion_tech}. It is defined as:
\begin{equation}
\label{eq4}
C_{S3}' = AW*C_{S3}=softmax(C_{S12}' * C_{S12}'^{T})*C_{S3}
\end{equation}
where $C_{S12}'=C_{S12}\oplus Linear(A^T)$. Similarly, \textbf{the (\{S2, S3\}$\rightarrow$ S1) case} takes the same direction as stated above except that now the source channels ($C_{S2}, C_{S3}$) first get up-scaled to match the dimensions of the target channels ($C_{S1}$).

\noindent \textbf{\{S1, S3\}$\rightarrow$ S2 fusion case.}
The first ($C_{S1}$) and third ($C_{S3}$) branch channels are down- and up-sampled respectively by $2\times$ to match the $S2$ dimensions, followed by their element-wise summation to generate $C_{S13}$. We apply the linear-layer on the audio embedding ($A^T$), which is separately added to each row of $C_{S13}$ via element-wise summation. The produced channels ($C_{13}'$) are used to obtain the attention-weights (AW) that get applied on target channels ($C_{S2}$) to yield audio-visual attended features ($C_{S2}'$) as shown in Fig. \ref{fig:fusion_tech}. We can define it as:
\begin{equation}
\label{eq5}
C_{S2}' = AW*C_{S2}=softmax(C_{S13}' * C_{S13}'^{T})*C_{S2}
\end{equation}
where $C_{S13}'=C_{S13}\oplus Linear(A^T)$.

\subsubsection{Visual Features Generation} 
The MSB higher-scales outputs ($C_{S1},C_{S2}$) are merged together with the lowest-branch output channels ($C_{S3}$) through ($3\times3$) convolution after down-scaling higher features via required average pooling ($AP$). The generated channels ($\in \R^{256*\frac{W}{32}*\frac{H}{32}})$ employ several convolution layers defined as follows: \{Conv2d(256,144,3,(1,1),1)-BN-ReLU, Conv2d (144,144,3,(4,1),1)-BN-ReLU\}. Where Conv2d (I,O,F,P,S) indicates I: input channels, O: output channels, F: F$\times$F filter, P: padding in (H,W), S: stride, and BN and ReLU denote Batch-Normalization \cite{ioffe2015batch} and ReLU \cite{nair2010rectified} activation function. The resultant channels ($\in \R^{Z*\frac{W}{128}*\frac{H}{72}})$ are reshaped to give the VFE module output as follows:
\begin{equation}
\label{eq6}
V=VFE(X),
\end{equation}
where $V\in \R^{P*Z}$, and $P\ (=\frac{W}{128}*\frac{H}{72})$ represents the total patches/regions in the input image. Intuitively, the $V$ matrix can be perceived as containing the Z-dimensional embedding for each image-patch, with $P$ total patches.

\begin{figure}[t]
	\begin{minipage}[b]{1\columnwidth}
		\begin{center}
			\centerline{\includegraphics[width=0.9\columnwidth]{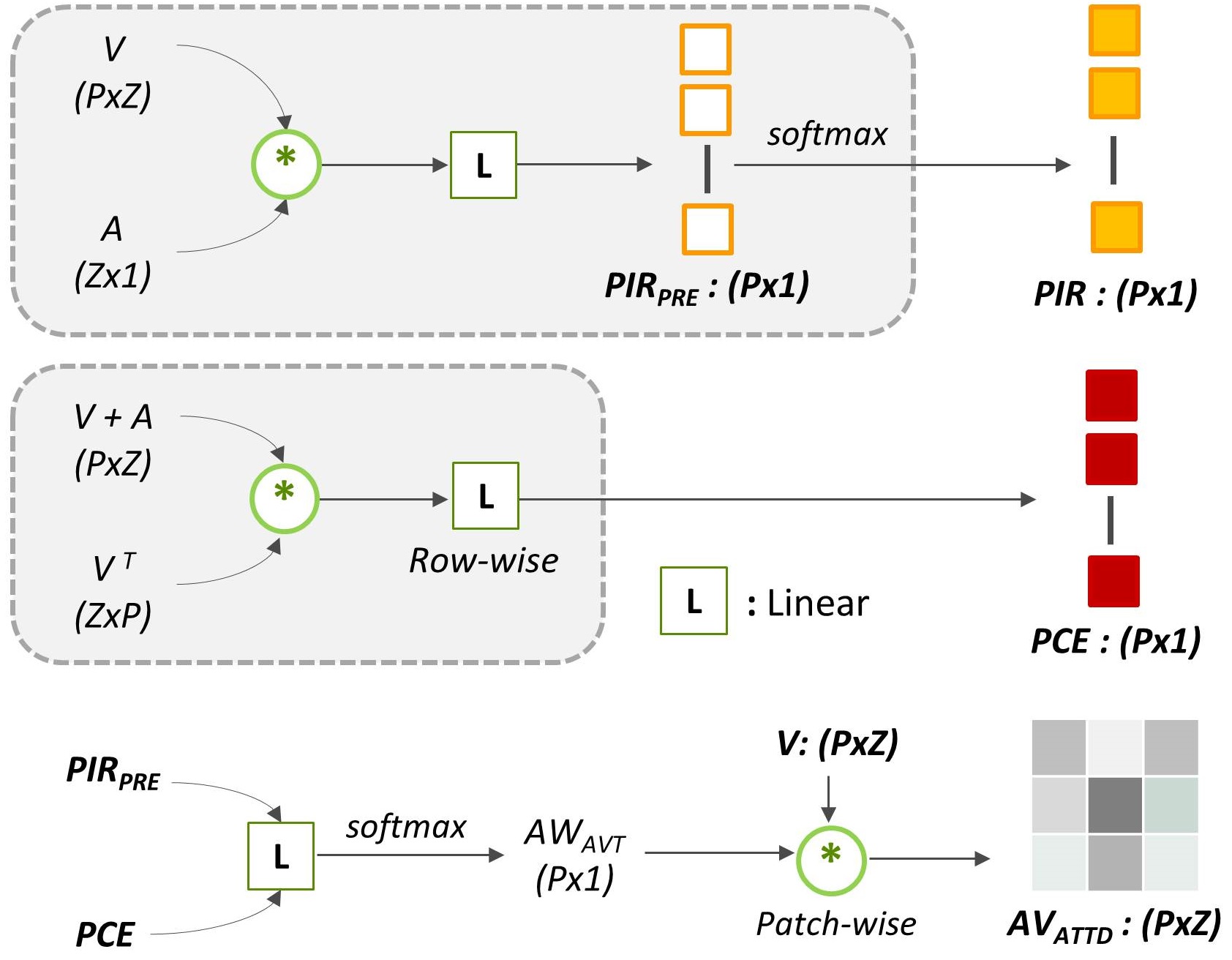}}
			\centerline{\footnotesize{(a) Audio-Visual Transformer (AVT) Unit.}}
		\end{center}
	\end{minipage}
   \begin{minipage}[b]{1\columnwidth}
		\begin{center}
			\centerline{\includegraphics[width=1\columnwidth]{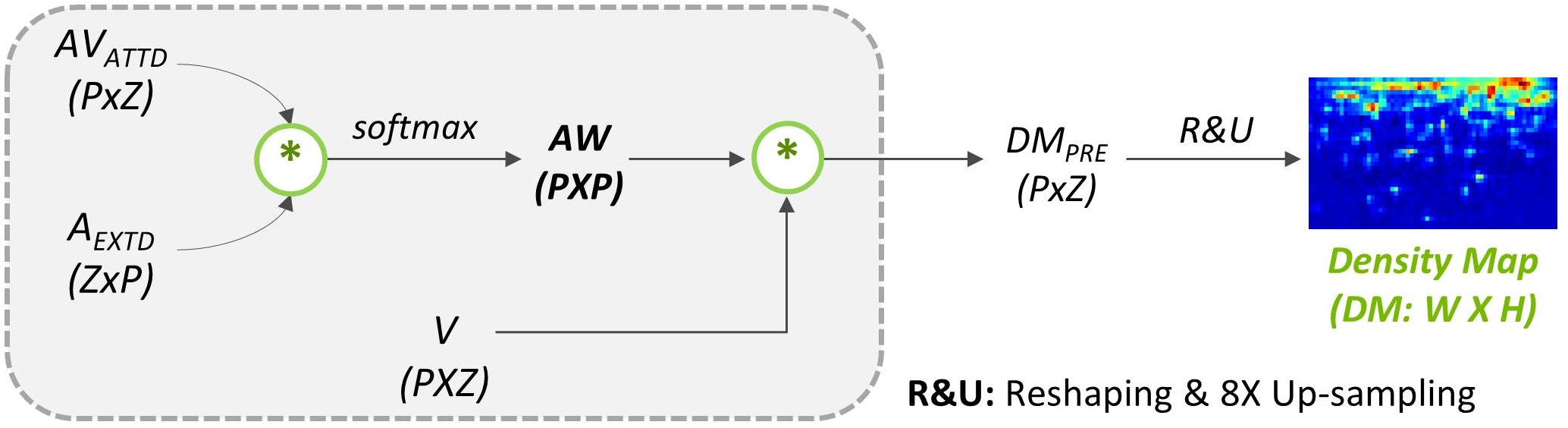}}
			\centerline{\footnotesize{(b) Cross-modality Co-attention Module (CCM).}}
		\end{center}
	\end{minipage}
	\vspace{-4mm}
	\caption{\footnotesize{Illustration of PIR, PCE, $AV_{ATTD}$, and DM computations.
	}}
    \vspace{-1mm}
    \label{fig:pir_pce}
\end{figure}

\subsection{Audio-Visual Transformer (AVT)} 
The purpose of the AVT module is twofold: 1) Calculate and output auxiliary Patch-Importance Ranking (PIR) and Patch-wise Crowd Estimate (PCE) Information, 2) Combine this information to generate third run-time modality to be used by the subsequent Co-attention ($CCM$) module. The AVT process, as shown in Fig. \ref{fig:pir_pce}(a), contains two separate streams to compute PIR and PCE. The AVT calculations are primarily inspired by the transformer-style dot-product attention amid using both visual ($V$) and audio ($A$) features. The PIR computation is defined as:
\begin{equation}
\label{eq7}
PIR=softmax(PIR_{PRE})
\end{equation}
where $PIR_{PRE}=Linear(V*A)$ and $PIR\in \R^{P*1}$. Intuitively, the PIR probability vector $ith$ value gives the percentage of total image people contained in the $ith$ image-patch. To set the ground-truth  PIR vector $jth$ value ($PIR_{GT}(j)$) for training, we use following formula:
\begin{equation}
\label{eq8}
PIR_{GT}(j)=\frac{CC_{GT}(j)}{CC_{GT}(image)}
\end{equation}
where $CC_{GT}(j)$ and $CC_{GT}(image)$ denote the actual crowd-count in the $jth$ patch and whole input image respectively. The KL-Divergence based loss function has been used to measure similarity between the PIR probability vector and the ground-truth probability distribution ($PIR_{GT}$):
\begin{equation}
\label{eq9}
Loss_{PIR}=\frac{1}{\sqrt{P}} \sum_{i=1}^{P} PIR_{GT}(i) \log(\frac{PIR_{GT}(i)}{PIR(i)})
\end{equation}

\noindent where $\frac{1}{\sqrt{P}}$ acts as a scaling factor. Similarly, the PCE vector is computed as:
\begin{equation}
\label{eq10}
PCE=Linear_{Row-wise}((V+A)* V^T)
\end{equation}
where $Linear_{Row-wise}$ indicates the row-wise linear-layer operation on the ($P\times P$) matrix to obtain ($PCE\in \R^{P*1}$) as shown in Fig. \ref{fig:pir_pce}(a). Intuitively, the $ith$ value in the PCE vector gives the network estimate for the $ith$ image patch. The ground-truth PCE vector computation strategy is the same as for PIR. The squared-normalized-difference loss function has been deployed for the PCE output, given as follows:
\begin{equation}
\label{eq11}
Loss_{PCE}=\sum_{i=1}^{P} (\frac{PCE_{GT}(i)-PCE(i)}{\sum_{j=1}^{P} PCE_{GT}(j)})^2
\end{equation}
where $PCE_{GT}$ indicates the ground-truth PCE vector and $\sum_{j=1}^{P} PCE_{GT}(j)$ denotes whole image actual people-count. The PIR and PCE information looks the same, but they invoke different yet relevant and effective behavior in the network because of different operational inputs being used for their calculation. In addition, the nature of both outputs differs as the PIR is probability-based, while the PCE directly regresses the crowd-count patch-wise. Next, the $PIR_{PRE}$ and $PCE$ pass through the linear-layer and softmax to produce the attention-weights ($AW_{AVT}$). The $AW_{AVT}$ is then applied on the original visual features ($V$) to give the PIR-PCE attended AVT output ($AV_{ATTD}\in \R^{P*Z}$), which acts as the third modality to be used in the next steps. This unique AVT strategy helps the network in focusing more on image regions with higher crowd-number and ignore the background patches. More importantly, the auxiliary mid-network PIR-PCE outputs aid both earlier and later-stage layers learning during the training process, and thus, resulting in significant improvement as demonstrated in experiments Sec. \ref{expi}.

\subsection{Cross-Modality Co-attention Module (CCM)} 
The co-attention module exploits the visual features ($V$) to perform the image-level crowd-estimation by jointly considering the audio features ($A$) and PIR-PCE attended channels ($AV_{ATTD}$). The transformer-inspired attention process is shown in Fig. \ref{fig:pir_pce}(b) and defined as:
\begin{equation}
\label{eq12}
DM_{PRE}=softmax(AV_{ATTD}*A_{EXTD})*V
\end{equation}
where $DM_{PRE}\in \R^{P*Z}$, and $A_{EXTD}$ is the ($Z\times P$) matrix containing $P$ times repeated vector $A$.

\setlength{\tabcolsep}{2.0pt}
\begin{table*}[t]\small

	\begin{center}
	\begin{tabular}{|c|c|c|c|c|c|c|c|c|c|c|c|c|c|c|c|}
    \hline

\multirow{3}{*}{Method} &   \multicolumn{2}{c|}{\multirow{2}{*}{Regular Images}} &\multicolumn{2}{c|}{Low Resolution} &  \multicolumn{4}{c|}{Gaussian Noise} & \multicolumn{4}{c|}{Low Illumination \& Gaussian Noise} & \multicolumn{2}{c|}{\multirow{2}{*}{Avg. Score}}\\ \cline{4-13}

&   \multicolumn{2}{c|}{} &\multicolumn{2}{c|}{$128\times 72$} &\multicolumn{2}{c|}{$\sigma = 25/255$} &\multicolumn{2}{c|}{$\sigma = 50/255$} &
\multicolumn{2}{c|}{R=0.2,B=25} &\multicolumn{2}{c|}{R=0.2,B=50} &  \multicolumn{2}{c|}{} \\  \cline{2-15}

 &  MAE & RMSE & MAE & RMSE  & MAE & RMSE & MAE & RMSE  & MAE & RMSE & MAE & RMSE & MAE & RMSE\\ \hline

MCNN \cite{zhang2016single} & 53.40   & 84.10 & 60.17 & 89.35  & 53.47 & 84.04 & 53.92 & 84.04  & 70.72 & 96.11 & 70.58 & 96.11 & 60.38 & 88.96 \\

CANNet \cite{liu2019context} & 15.41  & 28.96 & 22.16 & 39.60  & 13.31 & 27.23 & 14.20 & 28.04  & 26.03 & 49.11 & 33.14 & 58.27 & 20.71 & 38.54 \\

CSRNet \cite{li2018csrnet} & 13.88  & 28.79 & 17.14 & 30.64  & 13.79 & 28.01 & 14.55 & 29.15  & 35.78 & 62.76 & 45.88 & 75.40 & 23.50 & 42.46 \\

AudioCSRNet \cite{hu2020ambient} & 14.24  & 28.07 & 16.88 & 31.46  & 13.07 & 27.45 & 13.70 & 28.67  & 25.06 & 51.58 & 27.33 & 45.16 & 18.38 & 35.40 \\ \hline \hline

\textbf{CC-V (Ours)} & 12.97  & 25.76 & 16.91 & 32.82 & 13.31 & 28.79 & 13.92 & 29.01 & 26.03 & 55.72 & 27.59 & 58.69 & 18.46 & 38.47 \\

\textbf{CC-AV (Ours)} & \textbf{9.24}  & \textbf{19.81} & \textbf{11.18} & \textbf{26.25}  & \textbf{10.15} & \textbf{19.76} & \textbf{10.39} & \textbf{19.79}  & \textbf{20.14} & \textbf{44.58} & \textbf{21.17} & \textbf{40.86} & \textbf{13.71} & \textbf{28.51} \\ 

\textbf{CC-AV Boost (\%)} & \textbf{33.4}  & \textbf{29.4} & \textbf{33.8} & \textbf{14.3}  & \textbf{22.3} & \textbf{27.4} & \textbf{24.2} & \textbf{29.4}  & \textbf{19.6} & \textbf{9.2} & \textbf{22.5} & \textbf{9.5} & \textbf{25.4} & \textbf{19.5} \\ \hline

	\end{tabular}
	\end{center}
	   \vspace{-1mm}
	
	\caption{\footnotesize Quantitative Evaluation on the DISCO Benchmark \cite{hu2020ambient} based on regular and several low-quality images settings. (Here $R,B$ denote the hyper-parameters being used for the illumination decay-rate and Gaussian-noise standard deviation computations respectively as defined in \cite{hu2020ambient})}
 \vspace{-2mm}
 \label{table:quanti_results}
   
\end{table*}

\subsection{Final Crowd Estimate ($CE_{FINAL}$)} 
The $DM_{PRE}$ gets re-shaped and up-sampled $8\times$ to output the final crowd Density-Map ($DM\in \R^{W*H})$ as shown in Fig. \ref{fig:pir_pce}(b). We sum all $DM$ pixel-values to obtain the final crowd estimate ($CE_{FINAL}$) for the input image-audio. We deploy $L_{2}$-norm as the $DM$ loss-function, given as:
\begin{equation}
\label{eq13}
Loss_{DM}=\sum_{m=1}^{W}\sum_{n=1}^{H}(DM_{mn}-DM_{mn}')^2
\end{equation}
where $DM\in \R^{W*H}$, $DM'\in \R^{W*H}$ indicate estimated and ground-truth density-maps, respectively. The network total multi-task loss ($Loss_{TOTAL}$) will be as follows:
\begin{equation}
\label{eq14}
Loss_{TOTAL}=Loss_{PIR}+Loss_{PCE}+Loss_{DM}
\end{equation}

Unlike other existing audio-visual mechanisms \cite{hu2020ambient}, our scheme employs both global and local learning (inter-pixel and inter-patch)  in an explicit manner with the joint consideration of audio features, which empirically improves network performance significantly. It also helps in suppressing background regions at pixel, patch, and image-level.

\section{Training and Evaluation Details}
The only available audio-visual crowd counting dataset to-date (DISCO) \cite{hu2020ambient} contains images with the same $1920\times 1080$ resolution. As per convention, we resize them to $1024\times 576$ for better resources usage. Consequently, ($C_{S1},C_{S2},C_{S3}$) channels have ($32\times 256 \times 144$), ($64\times 128 \times 72$), ($128\times 64 \times 36$) dimensions respectively. Therefore, we have 64 image patches in total (i.e. $P=\frac{W}{128}*\frac{H}{72}=8*8=64$), and the value of $Z$ is set to 144. In case of low-resolution setting experiments, we have $128\times 72$ size input images as per the norm. During this setting, we train without any down-sampling in the initial convolution layers, giving dimensions of ($C_{S1},C_{S2},C_{S3}$) as ($32\times 128 \times 72$), ($64\times 64 \times 36$), ($128\times 32 \times 18$) respectively and $P=\frac{W}{32}*\frac{H}{18}=4*4=16$. 

To generate the ground-truth density map, we apply the $15\times15$ Gaussian kernel ($G \sim \mathcal{N}(0,\,4.0)\,$) on binary annotations, where the ground-truth annotations are available in terms of people head center location in the image. We employ Adam optimizer \cite{kingma2014adam} and the learning rate with an initial value of $1\mathrm{e}{-5}$ that decays by $0.99$ every epoch with total $500$ epochs. The training batch size is set to $4$ and model evaluation takes place after every epoch. To mitigate over-fitting, linear-layers are followed by the dropout layer with the drop-probability of $0.3$, and weight-decay ($\lambda = 1\mathrm{e}{-4}$) has been used.

\noindent \textbf{Evaluation Details.} We evaluate and compare our method with the state-of-the-art using standard evaluation metrics: Mean Absolute Error (MAE) and Root Mean Square Error (RMSE), defined as follows:
\begin{equation}
\label{eq15}
MAE = \frac{1}{N} \sum_{n=1}^{N} |E_{n}-C_{n}|, RMSE =\sqrt[]{ \frac{1}{N} \sum_{n=1}^{N} (E_{n}-C_{n})^{2}}
\end{equation}
where $C_{n}$ and $E_{n}$ indicate the ground-truth and estimated crowd for the test audio-image input $n$ respectively, and $N$ denotes the total test audio-image samples in the dataset.

\section{Experiments}
\label{expi}
We first discuss the numerical evaluation on the audio-visual and vision-only benchmark datasets, followed by the ablation study and visual analysis.

\subsection{Experiments on Audio-Visual Dataset}
\label{av_only}
DISCO \cite{hu2020ambient} is an up-to-date and only-available diverse audio-visual crowd dataset. It contains a total of $1,935$ high-resolution images ($1,920\times 1,080$) and corresponding one-second audio signals. We have $170,270$ people annotations in total with the minimum, maximum and average people per image equal to 1, 709, and 88, respectively. The (train, validation, test) split is pre-defined as ($1435$, $200$, $300$) respectively. We evaluate our network using both audio-visual and vision-only versions. Audio-visual (CC-AV) version is the same as discussed above, whereas the vision-only variant (CC-V) only uses the image input and is detailed in sub-section \ref{image_only}. As per the standard practice, we compare the proposed scheme with the state-of-the-art for three pre-defined image settings.

\begin{figure}[t]
	\begin{minipage}[b]{1.0\columnwidth}
		\begin{center}
			\centerline{\includegraphics[width=0.85\columnwidth]{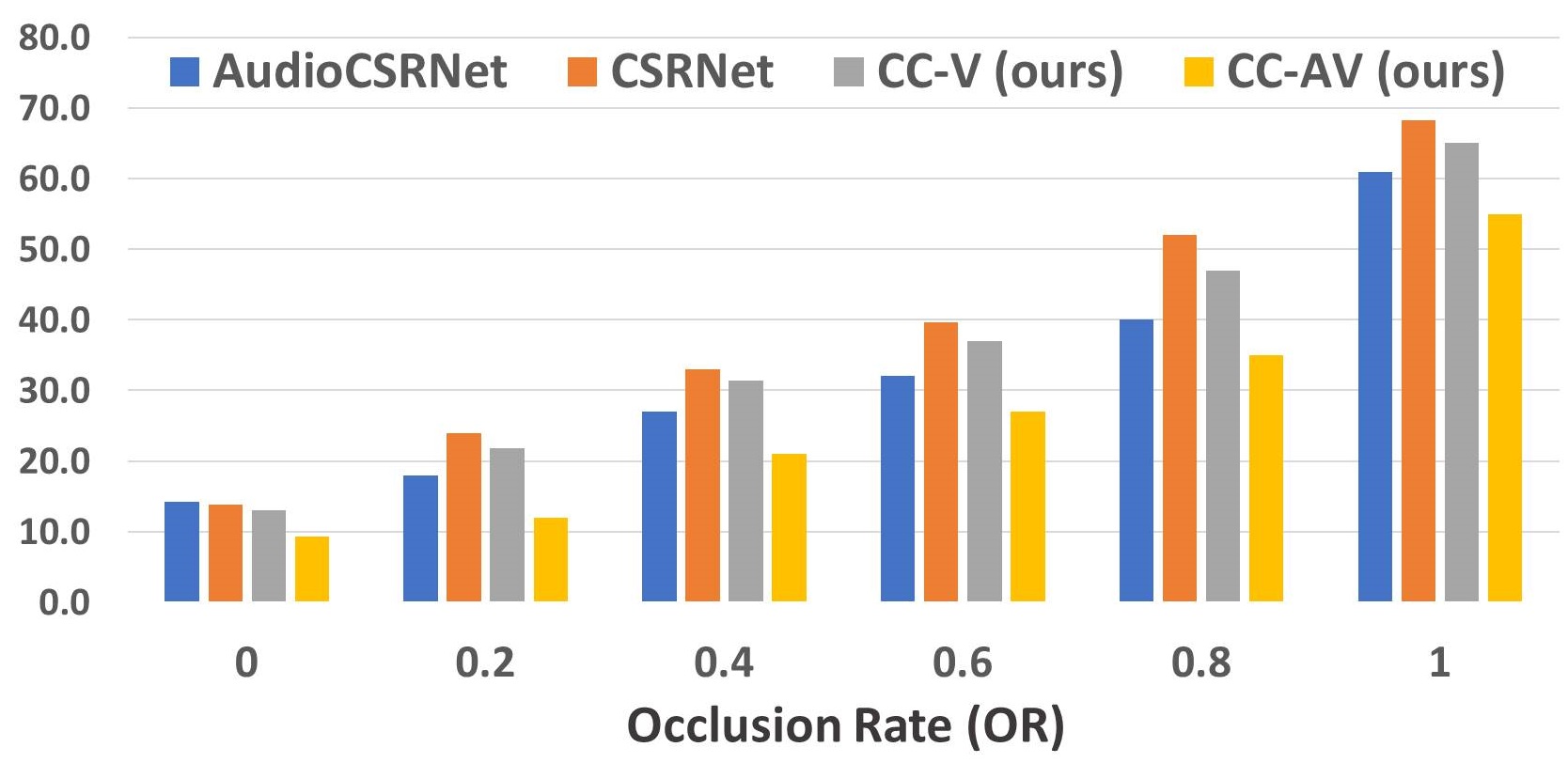}}
		\end{center}
	\end{minipage}
	\vspace{-6mm}
	\caption{\footnotesize{Occluded images setting based evaluation using MAE metric.}}
    \label{fig:graph_analysis}
    \vspace{-2mm}
\end{figure}

\noindent \textbf{Regular Images.} In this case, we use test images without any modification. The results, as shown in Table \ref{table:quanti_results}, indicate that both proposed network versions outperform the state-of-the-arts under all evaluation metrics with CC-AV giving $33.4\%$ and  $29.4\%$ error decrease for the MAE and RMSE metrics respectively. CC-AV performs significantly better than the CC-V, which directly implicates the benefit of including the audio modality.

\noindent \textbf{Low-Quality Images.} To check the robustness under severe conditions, we evaluate the model on three pre-defined standard settings: low-resolution, low-illumination, and strong noise. In the low-resolution setting, images are just $128\times 72$ in size. During the low-illumination study, random brightness reduction is followed by the Gaussian noise addition as defined in \cite{hu2020ambient}. Lastly, the Gaussian noise has been added in the strong noise case as given in \cite{hu2020ambient}. Observing the results for all three settings, as shown in Table \ref{table:quanti_results}, the proposed model (CC-AV) appears as the best choice with improvement up to $33.8\%$ and $29.4\%$ for MAE and RMSE respectively. The CC-V variant performance decreases in such extreme conditions because the visual information alone proves insufficient without any further aid.  

\noindent \textbf{Occluded Images.} In this setting, we occlude the image with a black rectangle using the Occlusion Rate (OR). The OR value lies in [0,1], meaning that image occlusion ranges from no occlusion ($OR=0$) to completely occluded ($OR=1$). The results, as shown in Fig. \ref{fig:graph_analysis}, show that the CC-AV model gives the best performance for the whole OR range as compared to the state-of-the-art methods (AudioCSRNet\cite{hu2020ambient} and CSRNet \cite{li2018csrnet}) on the MAE metric. All methods experience performance degradation as we increase the OR value due to the lack of more visual information. Our CC-V model yields a bigger error jump than CC-AV with the increase of OR values because it only relies on the visual information. Interestingly, in the case of no visual information (OR=1), CC-AV still performs better, indicating its robustness and better utilization of the audio-modality as compared to the best audio-visual models.

\begin{table}[t]\small

	
	\begin{center}
	\begin{tabular}{|c|c|c|c|c|}
    \hline

 & \multicolumn{2}{c|}{ShanghaiTech \cite{zhang2016single}} & \multicolumn{2}{c|}{UCF-QNRF \cite{idrees2018composition}}\\ \hline
Method & MAE  & RMSE & MAE  & RMSE\\ \hline
MCNN \cite{zhang2016single}  & 110.2  & 173.2 & 277   & 426   \\ 
Switch-CNN \cite{sam2017switching} & 90.4 & 135.0 & 228   & 445   \\ 
CSRNet \cite{li2018csrnet}   & 68.2  & 115.0 & -   & - \\  

CL\cite{idrees2018composition} & - & - & 132 & 191    \\ 

CAN \cite{liu2019context} & 62.3 & 100.0 & 107   & 183   \\ 

RRP \cite{chen2020relevant} & 63.2 & 105.7 & 93 & 156    \\

HA-CCN \cite{sindagi2019ha} & 62.9 & 94.9 & 118.1 & 180.4    \\ 

ADSCNet \cite{bai2020adaptive} & \textbf{55.4} & 97.7 & \textbf{71.3} & 132.5    \\

RPNet \cite{yang2020reverse} & 61.2 & 96.9 & - & -    \\

PRM-based\cite{sajid2020plug} & 67.8 & 86.2  & 94.5 & 141.9 \\ \hline \hline

\textbf{CC-V (Ours)}  & 58.7 & \textbf{81.3} & 75.4 & \textbf{125.6}   \\ \hline

	\end{tabular}
	\end{center}
	   \vspace{-1mm}
	   \caption{\footnotesize MAE and RMSE based evaluation  on image-only datasets.}

	\label{table:ST_results}
    \vspace{-3mm}
\end{table}

\subsection{Experiments on Image-only Datasets}
\label{image_only}
First, we discuss the design of the image-only variant (CC-V) of the proposed network. The CC-V structure remains the same as the CC-AV except that there is no available audio information ($A$) and thus the following changes have been made. 1) No $A$ based operation in the MSB three-branch fusion, PIR, PCE, and Co-attention processes. 2) Matrix operations required to compute $PIR_{PRE}$ have been replaced by the same set of operations being used for $PCE$. 3) Replace $A_{EXTD}$ with $V^T$ in the co-attention module. 

We compare our CC-V model on two image-only diverse benchmark datasets: UCF-QNRF \cite{idrees2018composition} and ShanghaiTech Part-A \cite{zhang2016single}. The UCF-QNRF dataset comprises of $1,535$ ($1,201$ train, $334$ test) images with total $1,251,642$ people annotations. On the other hand, ShanghaiTech dataset contains a diverse collection of $482$ crowd images ($300$ train, $182$ test). To avoid over-fitting in the case of ShanghaiTech dataset training, we use the model pre-trained on the UCF-QNRF benchmark, and train for only 250 epochs instead of 500. The images have been resized to $1,024\times 576$ with zero-padding if required. The results on both datasets are shown in Table \ref{table:ST_results}, where the proposed model CC-V yields the best performance for the RMSE metric ($5.2\%$ improvement for UCF-QNRF and $5.7\%$ for ShanghaiTech) amid producing reasonable results for the MAE as compared to the state-of-the-art schemes. These results demonstrate that the proposed scheme is also practical, robust, and highly effective in vision-only scenarios.

\begin{figure*}
\begin{minipage}[b][][b]{0.25\columnwidth}
		\begin{center}
			\centerline{\footnotesize{Regular}}
			\centerline{\footnotesize{Image case}}
			\centerline{\footnotesize{($1024\times 576$)}}
		\end{center}
	\end{minipage}	
\begin{minipage}[b][][b]{0.33\columnwidth}
		\begin{center}
			\centerline{\includegraphics[width=1.0\columnwidth]{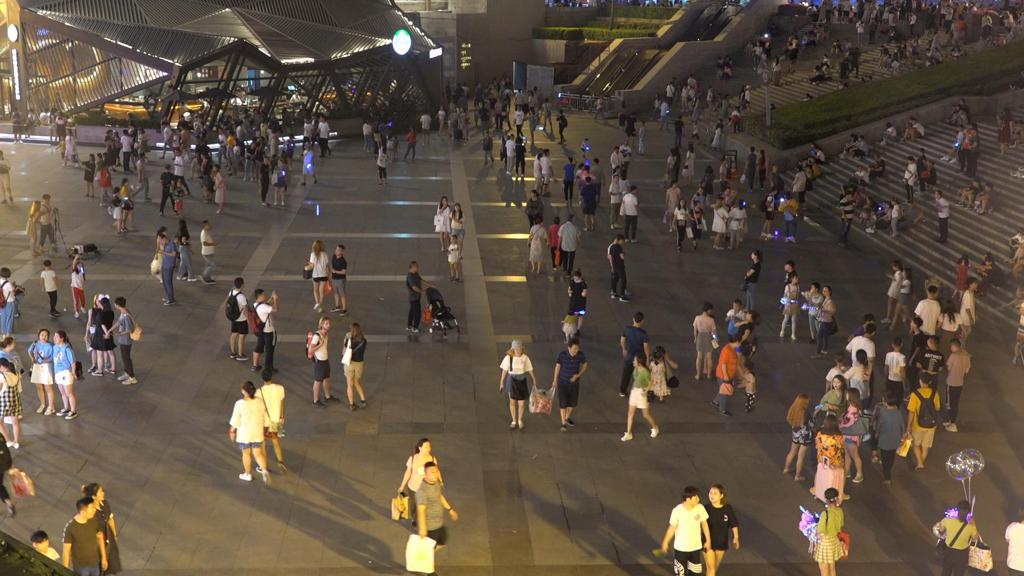}}
			
		\end{center}
	\end{minipage}	
	\begin{minipage}[b][][b]{0.125\columnwidth}
		\begin{center}
			\centerline{\includegraphics[width=1.0\columnwidth]{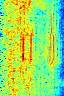}}
			
		\end{center}
	\end{minipage}
		\begin{minipage}[b][][b]{0.33\columnwidth}
		\begin{center}
			\centerline{\includegraphics[width=1.0\columnwidth]{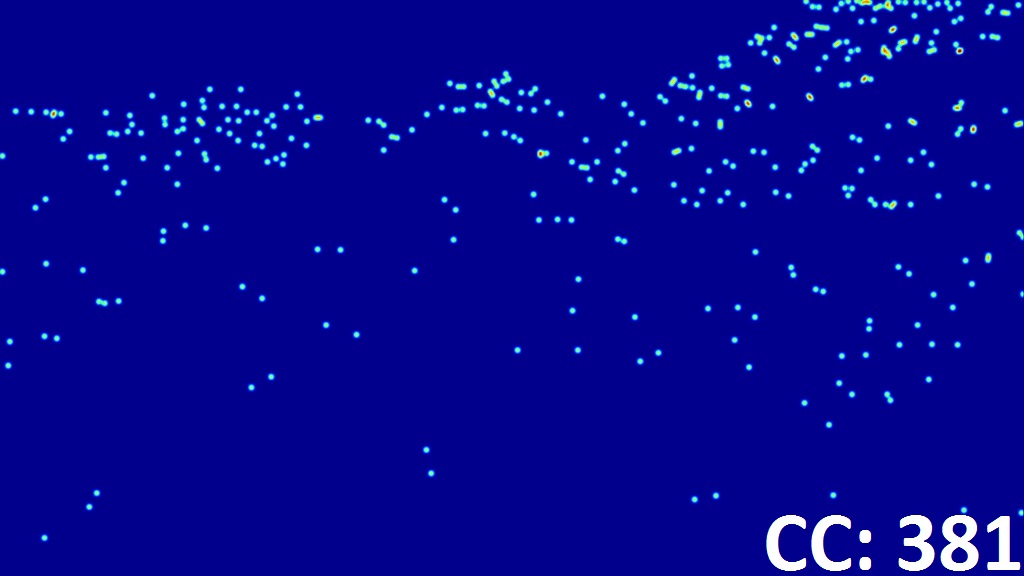}}
			
		\end{center}
	\end{minipage}
	\begin{minipage}[b][][b]{0.33\columnwidth}
		\begin{center}
			\centerline{\includegraphics[width=1.0\columnwidth]{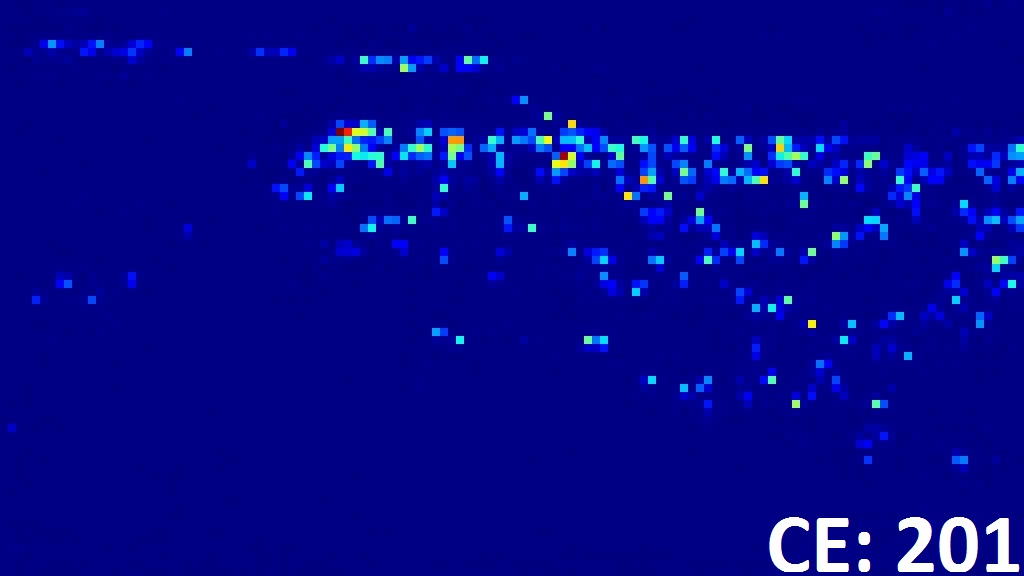}}
			
		\end{center}
	\end{minipage}
	\begin{minipage}[b][][b]{0.33\columnwidth}
		\begin{center}
			\centerline{\includegraphics[width=1.0\columnwidth]{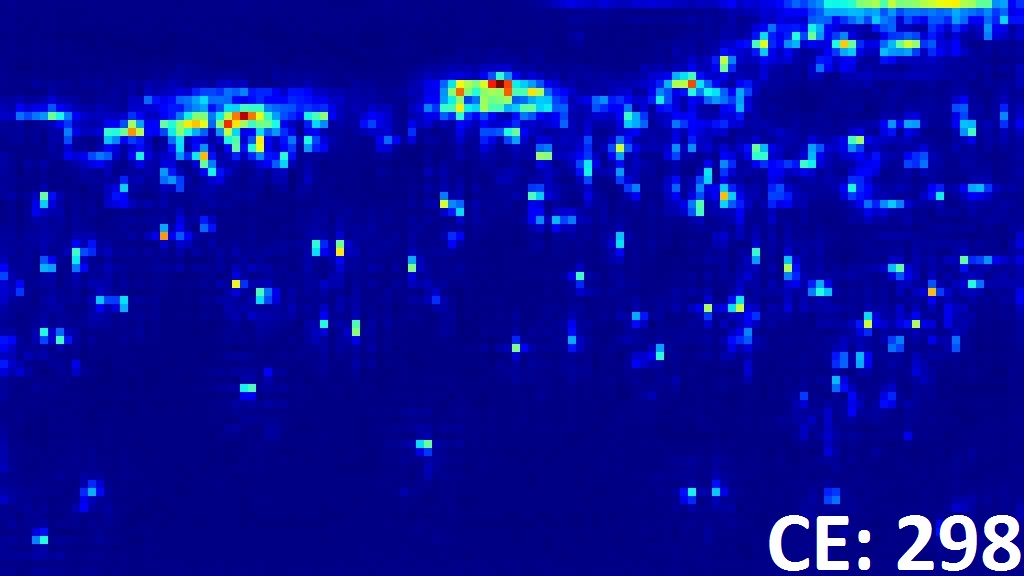}}
			
		\end{center}
	\end{minipage}
	\vspace{-1em}
	\begin{minipage}[b][][b]{0.33\columnwidth}
		\begin{center}
			\centerline{\includegraphics[width=1.0\columnwidth]{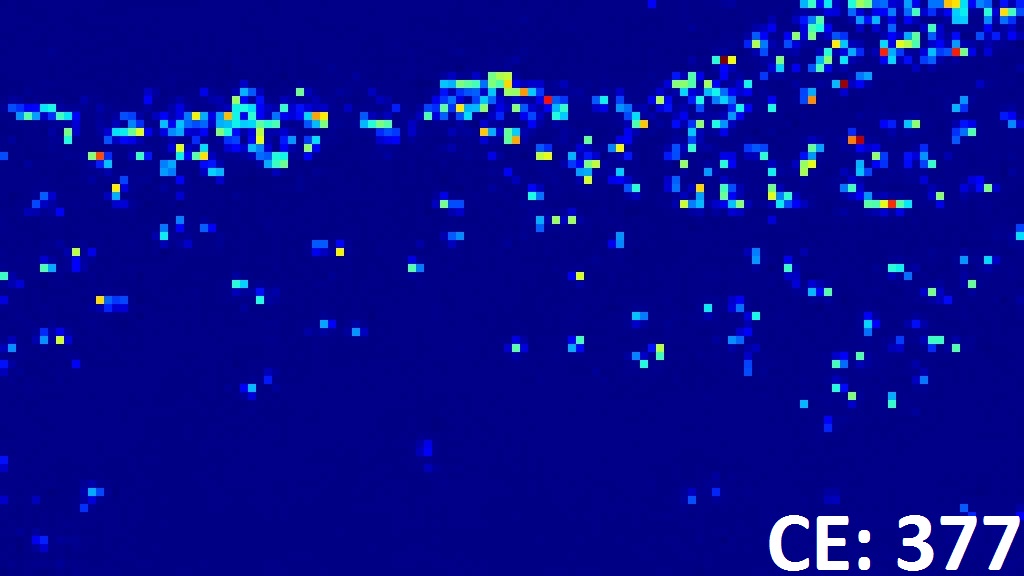}}
		\end{center}
	\end{minipage}
\vspace{-1em}
\begin{minipage}[b][][b]{0.25\columnwidth}
		\begin{center}
			\centerline{\footnotesize{Low-Resolution}}
			\centerline{\footnotesize{($128\times 72$)}}
		\end{center}
	\end{minipage}		
\begin{minipage}[b][][b]{0.33\columnwidth}
		\begin{center}
			\centerline{\includegraphics[width=1.0\columnwidth]{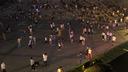}}
			
		\end{center}
	\end{minipage}	
	\begin{minipage}[b][][b]{0.125\columnwidth}
		\begin{center}
			\centerline{\includegraphics[width=1.0\columnwidth]{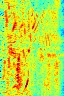}}
			
		\end{center}
	\end{minipage}
		\begin{minipage}[b][][b]{0.33\columnwidth}
		\begin{center}
			\centerline{\includegraphics[width=1.0\columnwidth]{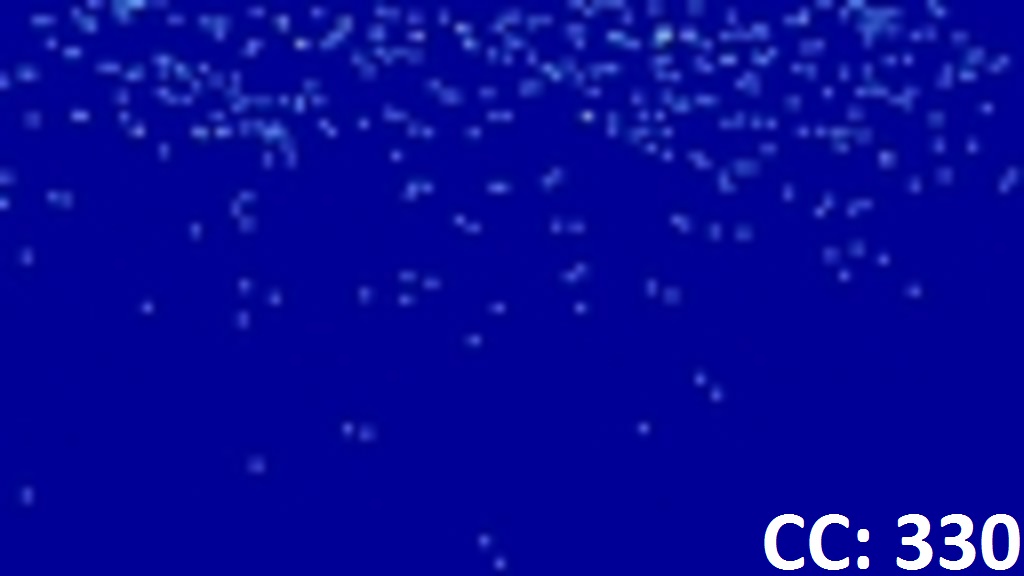}}
			
		\end{center}
	\end{minipage}
	\begin{minipage}[b][][b]{0.33\columnwidth}
		\begin{center}
			\centerline{\includegraphics[width=1.0\columnwidth]{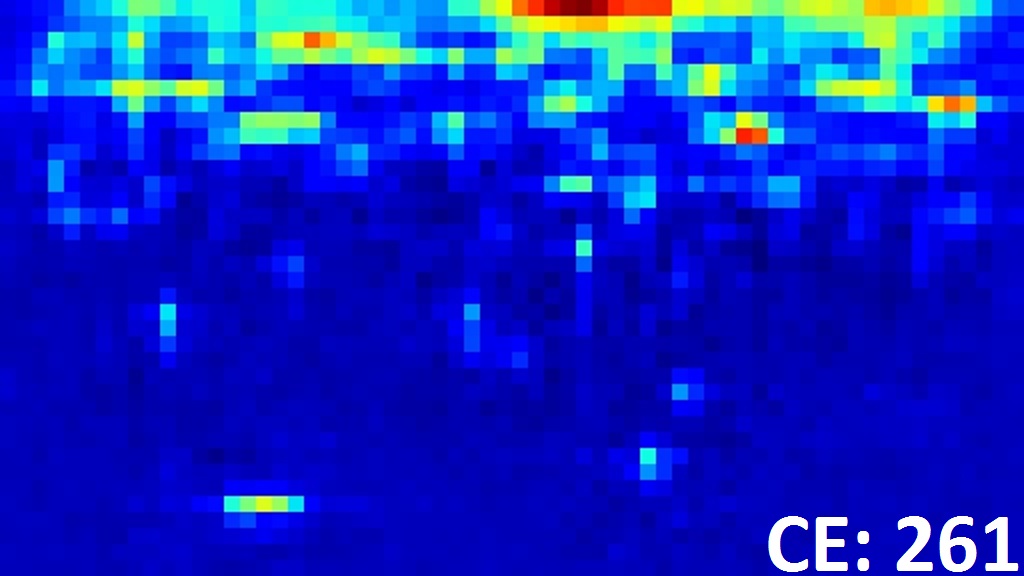}}
			
		\end{center}
	\end{minipage}
	\begin{minipage}[b][][b]{0.33\columnwidth}
		\begin{center}
			\centerline{\includegraphics[width=1.0\columnwidth]{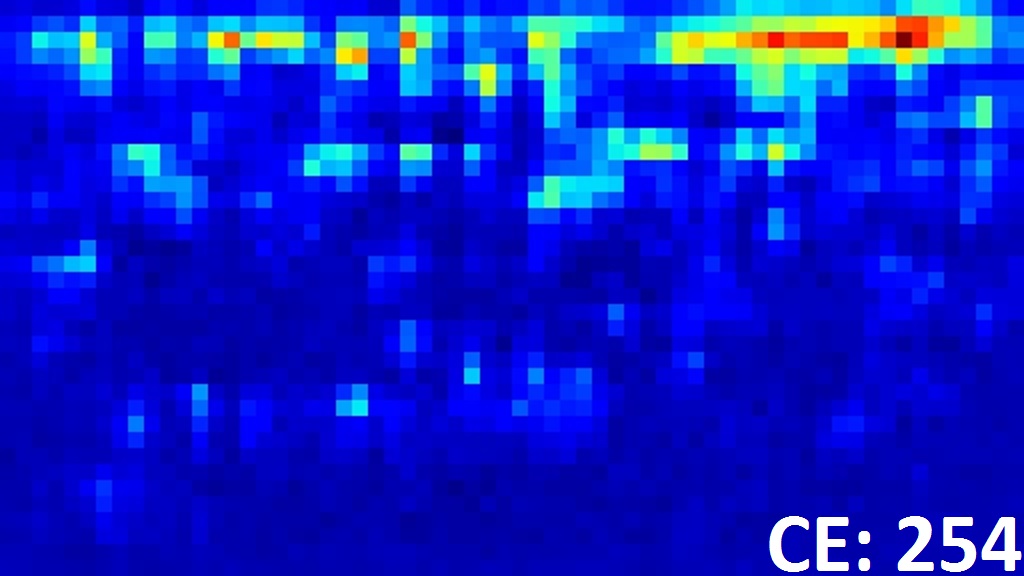}}
			
		\end{center}
	\end{minipage}
	\begin{minipage}[b][][b]{0.33\columnwidth}
		\begin{center}
			\centerline{\includegraphics[width=1.0\columnwidth]{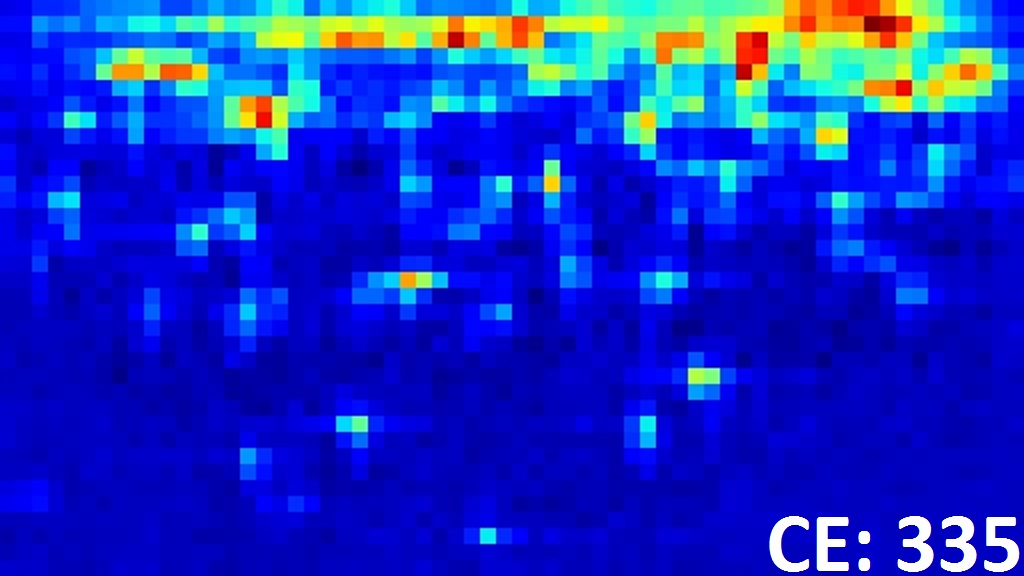}}
			
		\end{center}
	\end{minipage}
\vspace{-1em}
\begin{minipage}[b][][b]{0.25\columnwidth}
		\begin{center}
			\centerline{\footnotesize{Noisy case}}
			\centerline{\footnotesize{($\sigma = 50/255$)}}
		\end{center}
	\end{minipage}	
\begin{minipage}[b][][b]{0.33\columnwidth}
		\begin{center}
			\centerline{\includegraphics[width=1.0\columnwidth]{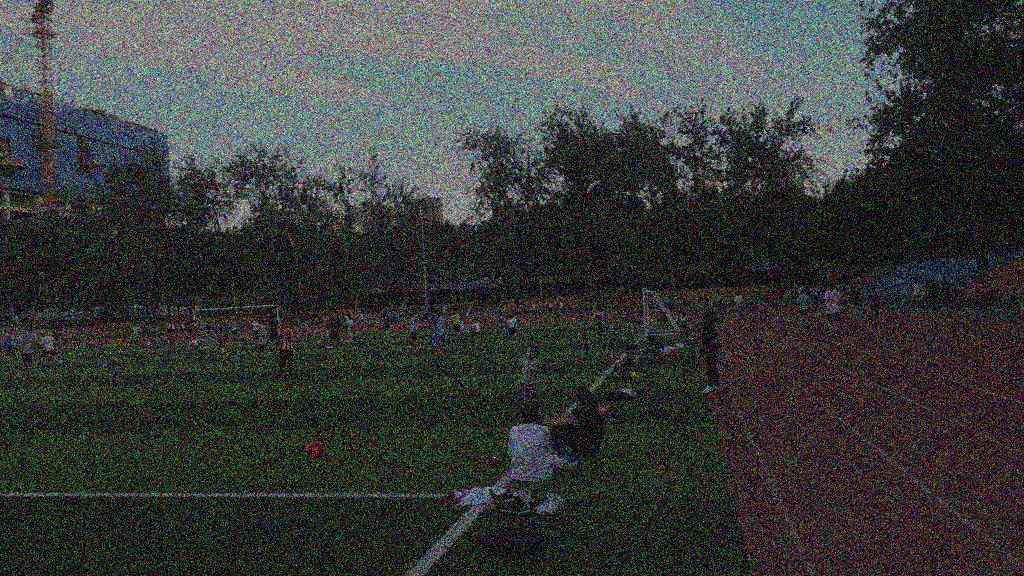}}
			
		\end{center}
	\end{minipage}	
	\begin{minipage}[b][][b]{0.125\columnwidth}
		\begin{center}
			\centerline{\includegraphics[width=1.0\columnwidth]{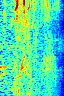}}
			
		\end{center}
	\end{minipage}
		\begin{minipage}[b][][b]{0.33\columnwidth}
		\begin{center}
			\centerline{\includegraphics[width=1.0\columnwidth]{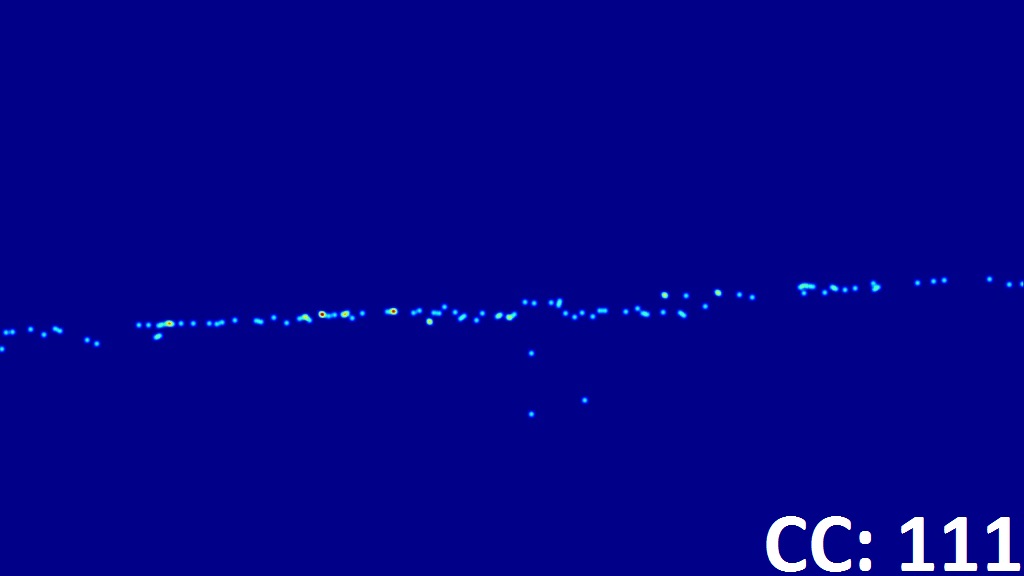}}
			
		\end{center}
	\end{minipage}
	\begin{minipage}[b][][b]{0.33\columnwidth}
		\begin{center}
			\centerline{\includegraphics[width=1.0\columnwidth]{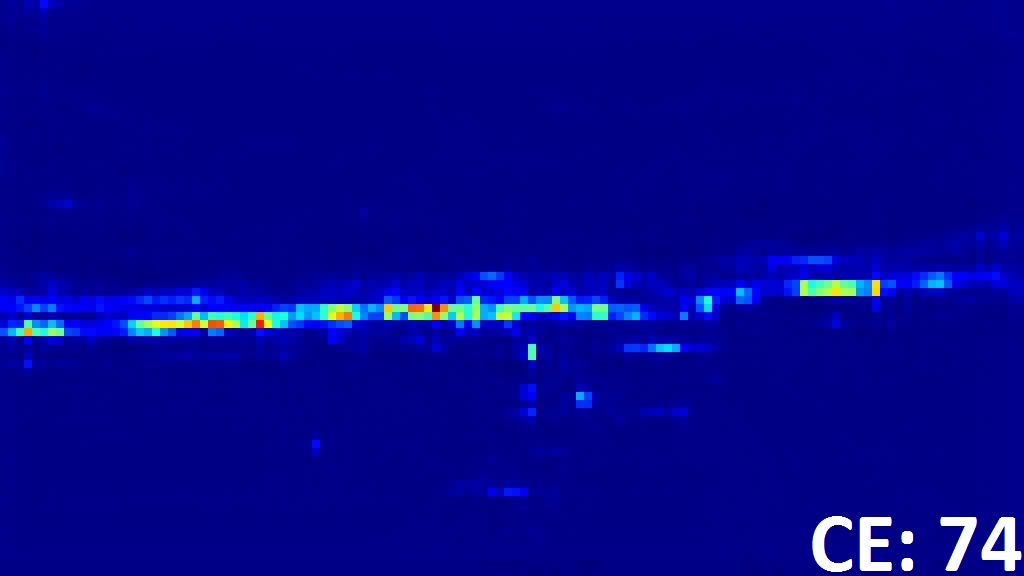}}
			
		\end{center}
	\end{minipage}
	\begin{minipage}[b][][b]{0.33\columnwidth}
		\begin{center}
			\centerline{\includegraphics[width=1.0\columnwidth]{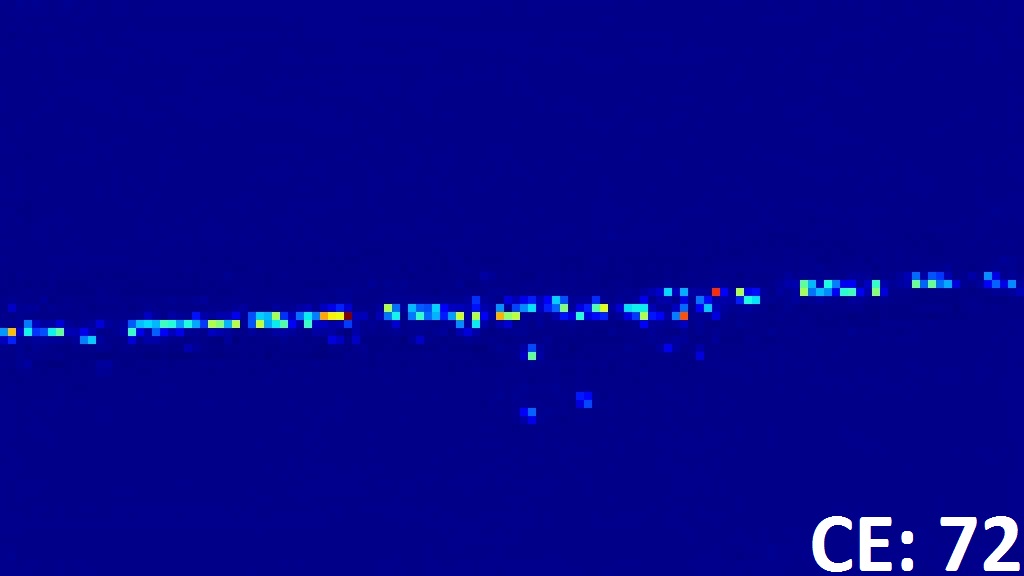}}
			
		\end{center}
	\end{minipage}
	\begin{minipage}[b][][b]{0.33\columnwidth}
		\begin{center}
			\centerline{\includegraphics[width=1.0\columnwidth]{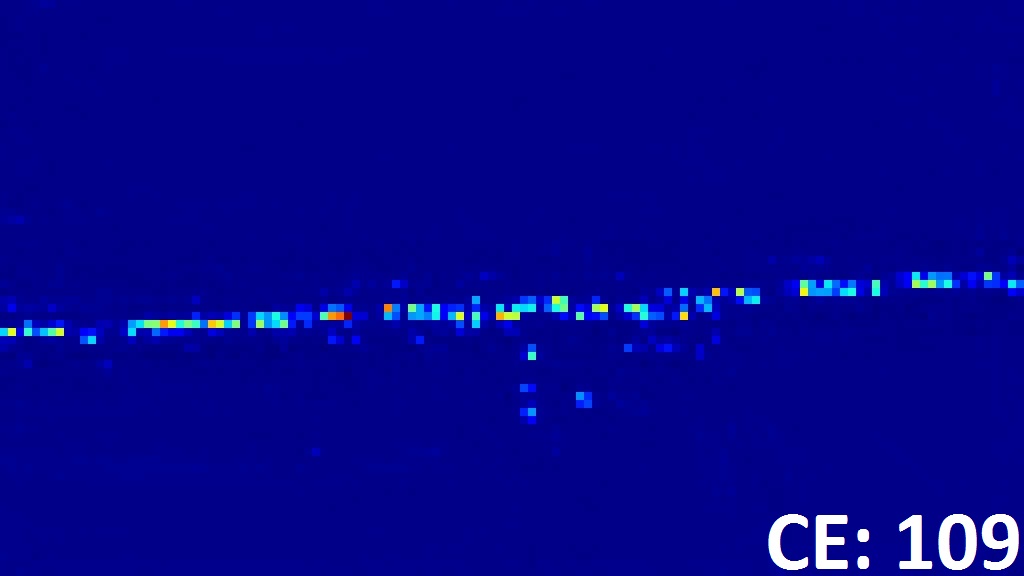}}
			
		\end{center}
	\end{minipage}
	
\begin{minipage}[b][][b]{0.25\columnwidth}
		\begin{center}
			\centerline{\footnotesize{$50\%$ Occluded}}
			\centerline{\footnotesize{Image}}
		\end{center}
	\end{minipage}	
\begin{minipage}[b][][b]{0.33\columnwidth}
		\begin{center}
			\centerline{\includegraphics[width=1.0\columnwidth]{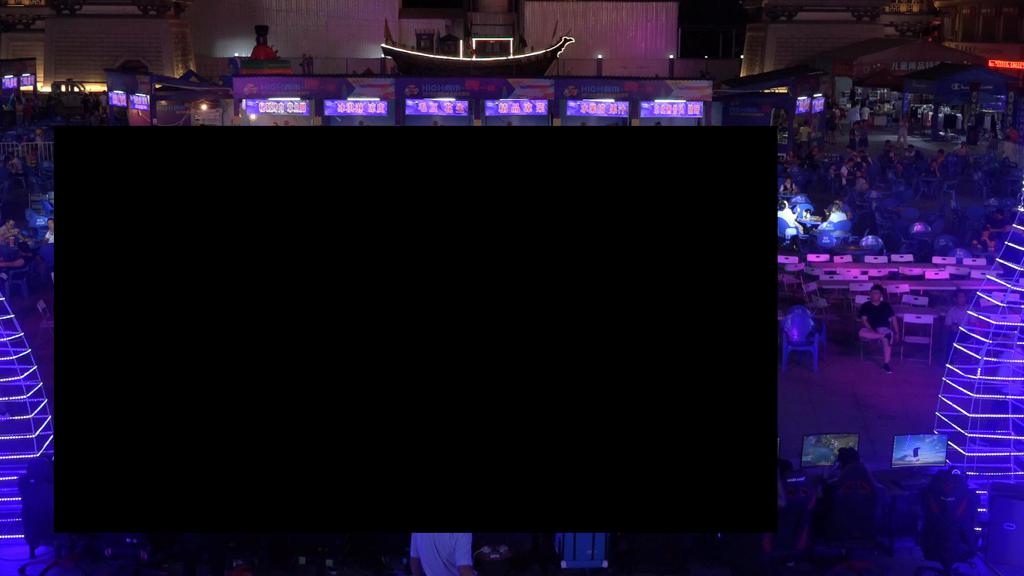}}
			
		\end{center}
	\end{minipage}	
	\begin{minipage}[b][][b]{0.125\columnwidth}
		\begin{center}
			\centerline{\includegraphics[width=1.0\columnwidth]{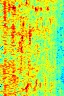}}
			
		\end{center}
	\end{minipage}
		\begin{minipage}[b][][b]{0.33\columnwidth}
		\begin{center}
			\centerline{\includegraphics[width=1.0\columnwidth]{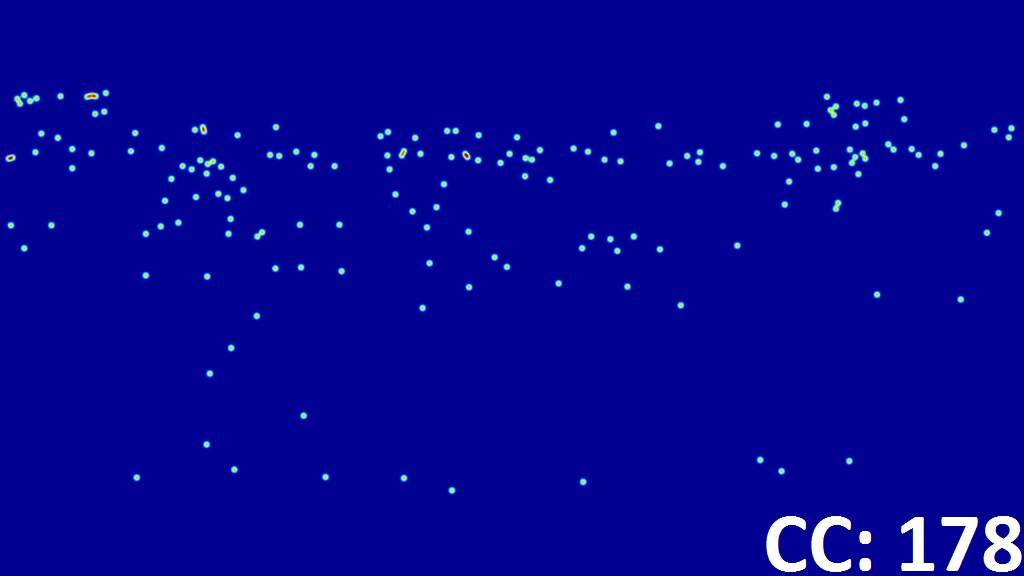}}
			
		\end{center}
	\end{minipage}
	\begin{minipage}[b][][b]{0.33\columnwidth}
		\begin{center}
			\centerline{\includegraphics[width=1.0\columnwidth]{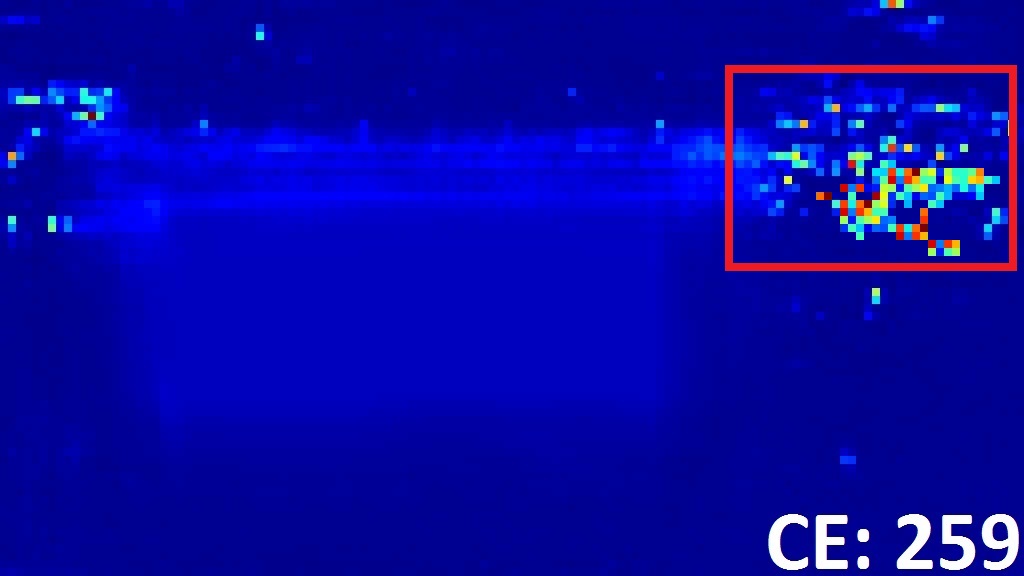}}
			
		\end{center}
	\end{minipage}
	\begin{minipage}[b][][b]{0.33\columnwidth}
		\begin{center}
			\centerline{\includegraphics[width=1.0\columnwidth]{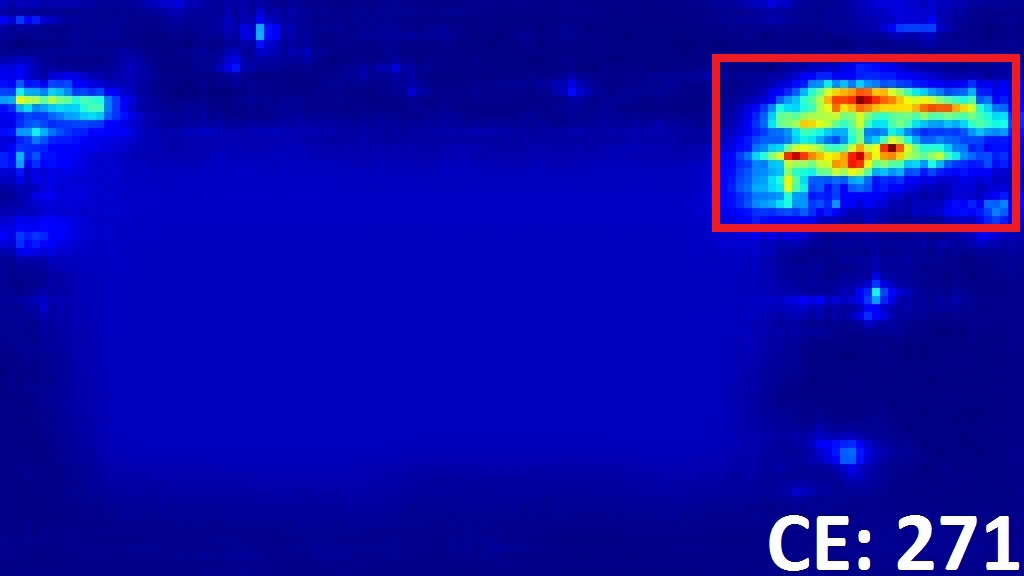}}
			
		\end{center}
	\end{minipage}
	\begin{minipage}[b][][b]{0.33\columnwidth}
		\begin{center}
			\centerline{\includegraphics[width=1.0\columnwidth]{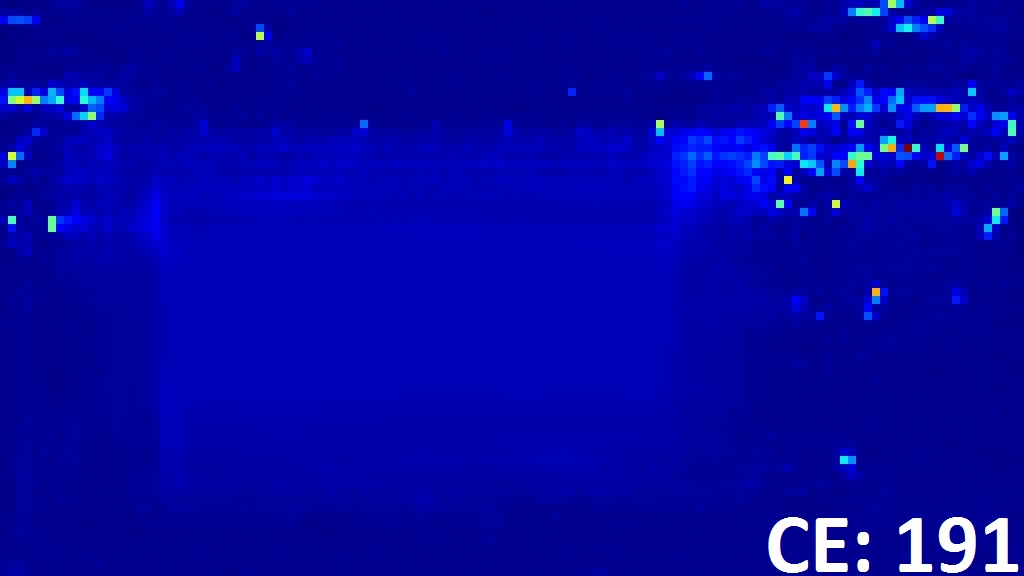}}
			
		\end{center}
	\end{minipage}

    \vspace{-2mm}
	\caption{\footnotesize{Ground truth (GT) density-map and crowd-count (CC) based qualitative comparison. (From Left to Right Column: Input Image, Audio Log Mel-Spectogram, GT density-map, AudioCSRNet network \cite{hu2020ambient} estimated density-map, CC-V model (ours) density-map, CC-AV (ours) density-map)}}
	\label{fig:qualityResults}
    \vspace{-2mm}
\end{figure*}

\subsection{Ablation Study}
In addition to the previous sub-section \ref{av_only} analysis on  audio-visual DISCO dataset various settings, here we further analyze and investigate the effect of different components on overall network performance during the following independent ablation studies.

\noindent \textbf{W/o explicit PIR, PCE.} No PIR vector output as well as no $Loss_{PRE}$ and $Loss_{PCE}$,  i.e. $Loss_{TOTAL}=Loss_{DM}$.

\noindent \textbf{W/o PIR or PCE branch in AVT unit.} In the first setting, we exclude the whole PIR computation stream and $Loss_{PIR}$, and only use the PCE stream and vector. In the second setting, we do vice versa by only keeping the PIR stream, and $Loss_{TOTAL}=Loss_{PIR}+Loss_{DM}$.

\noindent \textbf{W/o AVT.} No AVT module being deployed. Consequently, the CCM block uses $V$ instead of $AV_{ATTD}$. 

\noindent \textbf{W/o CCM.} No CCM module usage. The $AV_{ATTD}$ is considered as $DM_{PRE}$. 

\noindent \textbf{W/o $A^T$ in the MSB fusion.} No Audio information ($A$) has been used in any MSB three-branch fusion process.

\noindent \textbf{Using only single (S1) branch.} We only use one (S1) branch in the MSB multi-branch structure.

The results are listed in Table \ref{table:abl_results}, where we can observe that the (MAE,RMSE) errors increase by a noticeable margin in each case with as low as ($33.5\%$,$20.8\%$) and as high as ($51.1\%$,$50.1\%$) respectively. These evaluations indicate the effective importance of several network components including PIR, PCE, MSB, AVT, and CCM modules.

\setlength{\tabcolsep}{1.5pt}
\begin{table}[t]\small
\vspace{0mm}
	\begin{center}
	\begin{tabular}{|c|c|c|c|c|c|c|}
    \hline
 
 & & MAE & & RMSE \\   
 Ablation Setting  & MAE  & Increase & RMSE & Increase \\
 & &  (\%) & &  (\%)\\ \hline
 
 W/o explicit PIR, PCE & 16.0 & 42.3 & 28.2 & 29.8 \\ 
 W/o PIR branch in AVT & 16.7 & 44.7 & 27.6 & 28.2  \\
W/o PCE branch in AVT  & 15.5 & 40.4 &    28.1 & 29.5 \\ 
 W/o AVT & 18.9 & 51.1 & 39.7 & 50.1 \\ 
  
W/o CCM  & 17.3 & 46.6 & 31.4 & 36.9 \\ 
W/o $A^T$ in the MSB fusion  & 13.9  & 33.5 & 25.0 & 20.8  \\ 
W only S1 branch in MSB  & 14.3 & 35.4 & 26.8 & 26.1 
\\ \hline \hline
Default (CC-AV)  & \textbf{9.24} & -  & \textbf{19.81} & -  \\ \hline

	\end{tabular}
	\end{center}
	\vspace{0mm}
	
	\caption{\footnotesize Seven independent ablation studies on the effect of PIR, PCE, MSB, AVT and CCM components on the proposed network performance.}

	\label{table:abl_results}
    \vspace{0mm}
\end{table}

\subsection{Qualitative Analysis}
We present a few visual results as shown in Fig \ref{fig:qualityResults}. These results contain both regular (top row) and low-quality image cases (last three rows). For each input image, we display the input image, Log Mel-Spectrogram (LMS), ground-truth crowd density-map and count (CC) as well as predicted density-map and crowd-estimate (CE) being generated by our CC-AV, CC-V networks, and state-of-the-art AudioCSRNet \cite{hu2020ambient}. We can easily observe that the proposed audio-visual model (CC-AV) yields the most effective and fine-grained results as compared to the visual-only variant (CC-V) and AudioCSRNet \cite{hu2020ambient} in both regular and low-quality cases. However, the CC-V model experiences more error increase in low-quality cases due to lack of audio modality. These results also demonstrate that the proposed CC-AV network has significantly improved performance because of the better inclusion of the audio modality. Interestingly, the CC-AV performance is naturally better for regular images as visual information fades away in low-quality cases. One mentionable case is that of $50\%$ random image occlusion (last row of Fig. \ref{fig:qualityResults}). CC-V highly over-estimates in the non-occluded regions (highlighted in the red rectangular area) to compensate for the occluded area, and lacks the audio-modality aid to better estimate for the hidden region. Similarly, AudioCSRNet \cite{hu2020ambient}
also over-estimates in the same manner due to under-utilization of the audio information. On the other hand, our CC-AV model performs more robustly 
for both occluded and non-occluded regions.

\vspace{-0.4cm}
\section{Conclusion}
\vspace{-0.3cm}
In this paper, we have presented a new audio-visual multi-task network for effective people counting
by introducing explicit PIR and PCE information for better modalities association, and also 
producing a third run-time modality. This modality greatly helps the cross-modality fusion process to yield a better crowd estimate. We have also deployed a unique multi-branch structure to extract rich visual features and also proposed the image-only variant of our model. Experimental evaluation on standard benchmarks reveals the superior performance of our networks.


\balance
{\small
\bibliographystyle{ieee_fullname}
\bibliography{main}

}

\end{document}